\documentclass{article}

\usepackage[preprint]{neurips_2026}

\usepackage{hyperref}

\usepackage{algorithm}
\usepackage{algpseudocode}
 
\usepackage[utf8]{inputenc} 
\usepackage[T1]{fontenc}    
\usepackage{hyperref}       
\usepackage{url}            
\usepackage{booktabs}       
\usepackage{amsfonts}       
\usepackage{nicefrac}       
\usepackage{microtype}      
\usepackage{xcolor}         
\usepackage{wrapfig}

\usepackage{amsmath}
\usepackage{amssymb}
\usepackage{mathtools}
\usepackage{amsthm}
\usepackage{arydshln}

\usepackage[capitalize,noabbrev]{cleveref}

\usepackage{amsmath,amsfonts,bm}

\def\eqref#1{equation~\ref{#1}}

\def\1{\bm{1}}

\def\vb{{\bm{b}}}
\def\vc{{\bm{c}}}

\def\vq{{\bm{q}}}

\def\vu{{\bm{u}}}

\def\vw{{\bm{w}}}

\def\mA{{\bm{A}}}

\def\mI{{\bm{I}}}

\DeclareMathAlphabet{\mathsfit}{\encodingdefault}{\sfdefault}{m}{sl}
\SetMathAlphabet{\mathsfit}{bold}{\encodingdefault}{\sfdefault}{bx}{n}

\newcommand{\R}{\mathbb{R}}

\theoremstyle{plain}

\theoremstyle{definition}

\theoremstyle{remark}

\usepackage[textsize=tiny]{todonotes}
\usepackage{xcolor}

\definecolor{mygray}{gray}{0.9}

\title{Leech Lattice Vector Quantization\\for Efficient LLM Compression}

\author{
  Tycho F. A. van der Ouderaa \quad
  Mart van Baalen \quad
  Paul Whatmough \quad
  Markus Nagel \\
  Qualcomm AI Research\thanks{
    Qualcomm AI Research is an initiative of Qualcomm Technologies, Inc. \\
    Correspondence to: \texttt{\{touderaa,mart,pwhatmou,markusn\}@qti.qualcomm.com}.\vspace{-2em}
  }
}

\begin{document}

\maketitle

\begin{abstract}
Scalar quantization of large language models (LLMs) is fundamentally limited by information-theoretic bounds. While vector quantization (VQ) overcomes these limits by encoding blocks of parameters jointly,  practical implementations must avoid the need for expensive lookup mechanisms or other explicit codebook storage. Lattice approaches address this through highly structured and dense packing.
This paper explores the Leech lattice, which, with its optimal sphere packing and kissing configurations at 24 dimensions, is the highest dimensional lattice known with such optimal properties. To make the Leech lattice usable for LLM quantization, we extend an existing search algorithm based on the extended Golay code construction, to i) support indexing, enabling conversion to and from bitstrings without materializing the codebook, ii) allow angular search over union of Leech lattice shells, iii) propose fully-parallelisable dequantization kernel. Lastly, we provide a geometric reinterpretation of combining shape--gain quantization with GPTQ-style Hessian corrections: the standard scale-correction step of shape--gain acts as a \emph{retraction} onto a product of spheres, yielding a \emph{Spherical GPTQ} primarily acting on directions. We find that low-angular-distortion LLVQ reduces sensitivity to Hadamard/rotation preprocessing, and enables a strong \emph{Hadamard-free} PTQ in practice. LLVQ delivers state-of-the-art LLM quantization performance, outperforming recent methods such as Quip\#, QTIP, and PVQ. The results highlight the effectiveness of high-dimensional lattices for scalable, theoretically grounded model compression.
\end{abstract}

\section{Introduction}
\label{introduction}

Quantization is a critical technique for compressing large language models (LLMs). Traditionally, this has been approached through scalar quantization, where individual weights are represented using fewer bits. While simple and widely adopted, classical results in rate–distortion theory (originating with Shannon) show that memoryless mappings are, in general, suboptimal: achieving the optimal distortion at a given rate typically requires coding over blocks rather than symbol-by-symbol mappings \citep{shannon1948mathematical,shannon1959coding}. Perhaps surprisingly, this even holds for completely independent and isotropically distributed sources, such as Gaussian vectors, where block coding strictly outperforms scalar methods in the rate-distortion trade-off \citep{gray2002quantization}. Consequently, scalar quantization is fundamentally limited when targeting aggressive compression without significant accuracy loss.

To overcome these limitations, vector quantization (VQ) \citep{gersho2012vector} encodes \emph{blocks} of weights jointly. Concretely, a $d$‑dimensional block represented by a $b$‑bit index selects one codeword from a set of size $2^{b}$, yielding an average rate of $b/d$ bits per weight. From a deep‑learning practitioner’s perspective, this is akin to assigning a dedicated \emph{dtype} to an entire block of weights rather than to each scalar entry: the block is stored as a single compact integer index instead of many independent scalars. A \emph{naive} realization of this idea is to materialize the codebook explicitly and perform nearest‑neighbor lookup among its $2^{b}$ high‑dimensional codewords. \emph{GPTVQ} \citep{van2024gptvq} demonstrates that such unstructured VQ can be applied to LLMs; however, the explicit‑table approach scales poorly with dimension $d$, because both storage and lookup costs grow exponentially with the vector dimensionality.

This underscores a limitation of the classical theory: Shannon’s results establish the optimality of block coding in principle, yet offer no guidance on practical implementations. The key challenge, therefore, is to design VQ schemes that avoid an explicitly stored codebook and exhaustive nearest-neighbor search, while still admitting large representable sets. A considerable body of research has explored how to impose structure on vector quantizers to avoid the prohibitive cost of unstructured nearest‑neighbor search. Recent work on LLM quantization have exploited such structured representations, such as Quip\# \citep{tseng2024quip}, which uses the $E_8$ lattice; QTIP \citep{tseng2024qtip}, which employs trellis-based constructions to scale to higher dimensions; and PVQ \citep{van2024pyramid}, which uses flexible high-dimensional pyramids as quantization rules.

The use of lattices for quantizing LLM weights was recently popularized by QuIP\#, which employs the $E_8$ lattice in eight dimensions. Together with the Leech lattice, these occupy a distinguished place in mathematics: $E_8$ achieves optimal sphere packing in dimension~8, while the Leech lattice does so in dimension~24. These are the highest dimensions in which optimal lattice packings are known, proven only recently through breakthroughs in harmonic analysis and modular forms, work that earned Maryna Viazovska the 2022 Fields Medal \citep{IMU_FieldsMedals_2022}.


Practical scalable VQ methods must deliver good rate–distortion performance on target distributions while enabling fast quantization and dequantization. To do so, it should support: (i) efficient nearest‑neighbour search on the (implicit) quantization grid, (ii) a compact integer or bitstring representation of each quantized vector, and (iii) a fast mapping from this index back to its corresponding representative vector. All without explicitly storing the codebook in memory. Lastly, we find that matching the original norm in shape--gain can be viewed as a \emph{retraction} onto a constant-norm manifold, which when paired with GPTQ, suggests a natural \emph{Spherical GPTQ} interpretation in which error-feedback updates act primarily on directions rather than magnitudes.

Our proposed method, Leech Lattice Vector Quantization (LLVQ), a codebook‑free quantization framework built on the structure of the Leech lattice, satisfies all these criteria. LLVQ leverages the geometric structure of the Leech lattice to provide state-of-the-art vector quantization for large language models, delivering superior accuracy–model‑size trade-offs. Specifically, our contributions are as follows:
\begin{enumerate}
    \vspace{-0.4em}
    \item Extend codebook-free nearest neighbour algorithm of \citet{adoul1988nearest} on the Leech lattice to allow \textit{indexing}, enabling conversion to and from indices/bitstrings without materializing a codebook, required for actual compressed representations.
    \vspace{-0.4em}
    \item Extend codebook-free nearest neighbour algorithm on the Leech lattice to allow \textit{angular search over union of Leech lattice shells}, required for shape-gain Leech lattice quantization.
    \vspace{-0.4em}
    \item Propose a fully parallelizable kernel for fast dequantization of spherically bounded Leech lattice points using fast modulo arithmetic.
    \vspace{-0.4em}
    \item Spherical GPTQ: We show that the standard scale correction used in shape--gain quantization, when interleaved with GPTQ-style Hessian error feedback, is naturally interpreted as a (hyper)sphere-constrained variant of GPTQ on a product of spheres. We find that this is very effective for directional LLVQ codes, and allows for a strong \emph{Hadamard-free} PTQ.
    \vspace{-0.4em}
    \item Scientific findings on Gaussian source: establish that union of shells achieves lower angular distortion than using single shells (\autoref{app:union-of-shells}), and demonstrate that Leech shape-gain codes can improve signal-to-noise over spherical shaping (\autoref{app:spherical-bounding-vs-shape-gain}).
    
\end{enumerate}
LLVQ achieves state‑of‑the‑art \textbf{2 bit per weight} PTQ quantization of large language models, consistently surpassing AQLM, QuIP\# and QTIP across perplexity metrics and downstream tasks on various common LLM models, such as from the Llama-2, Llama-3, Ministral-3 and Qwen-v3 architectures. In addition, the algorithm naturally allow a wide variety of bit-widths (unlike competing approaches, often relyng on techniques such as residual vector quantization RVQ \citep{tseng2024quip} to increase bitrates). Further  paired with Spherical GPTQ, LLVQ remains competitive even without Hadamard rotations, suggesting a practical route to rotation-free, low-latency PTQ. Altogether, the findings highlight high‑dimensional lattice methods as a powerful path toward scalable, low‑distortion compression of modern neural networks. 

\section{The Leech Lattice}

In this work, we use the \emph{Leech lattice} \(\Lambda_{24}\) as the foundation for our quantization codebook. Its exceptional symmetry, high minimum distance, and rich shell structure make it uniquely suited for constructing efficient, low-distortion spherical codes with fast encoding and decoding procedures.

\subsection{Definition and Properties}

A lattice in \(\mathbb{R}^n\) is a discrete additive subgroup generated by an integer linear combination of basis vectors \(b_1,\dots,b_n\):
$\Lambda = \left\{ \sum_{i=1}^n k_i b_i \,\middle|\, k_i\in\mathbb{Z} \right\}.$
\noindent The Leech lattice \(\Lambda_{24}\) is a renowned 24-dimensional lattice, owing to its numerous optimal geometric properties. It achieves the densest and provably optimal sphere packing in 24 dimensions and exhibits a massive automorphism group (of size roughly \(8.3\times 10^{18}\)), reflecting a high degree of symmetry. Dense sphere packing is often cited as a desirable property for quantization because, under the standard high-rate assumption that the source is approximately uniform at the scale of the Voronoi cell. In addition, the Voronoi cell of \(\Lambda_{24}\) has an exceptionally low normalized second moment, further minimizing distortion under these second-order (quadratic) approximations. We note that normalizing Leech lattice vectors also produce remarkably uniform spherical codes, allowing high-performant finite bitrate quantization not just in Euclidean, but also in spherical geometry (e.g. shape-gain quantization, see \autoref{sec:finite-codes-from-lattices}).

Several equivalent constructions of $\Lambda_{24}$ exist (see \citealp{conway2013sphere} for an authoritative reference). We adopt the formulation based on the extended binary Golay code, which provides an explicit coordinate representation of the lattice. The proposed neighbour search and indexing procedures operate directly in $\mathbb{R}^{24}$ while internally exploiting the Golay-derived structure. This avoids having to enumerate or materialize an astronomically large set of lattice points.


\subsection{Finite Codes from Lattice Shells}
\label{sec:finite-codes-from-lattices}

Lattices are infinite, highly structured point sets in $\mathbb{R}^n$. While they provide a rich mathematical structure, quantization requires a finite subset, where the codebook must consist of a limited number of points so that each codeword can be represented using a fixed number of bits (just as integers form an infinite set, but any integer-based datatype must restrict its representable range). To transform a lattice into a practical finite representation, it is common to use \textit{spherical shaping}, in which we retain only those lattice points whose Euclidean norm does not exceed a chosen radius.

A useful property of many lattices, including the Leech lattice, is that their points naturally partition into \emph{shells}: sets of lattice points lying at the same squared Euclidean norm. Consequently, these shells occur at discrete radii (of integer squared norm) and exhibit rich combinatorial structure that can be exploited for fast search and efficient enumeration. We now formalize the partitioning of the Leech lattice into shells. For each integer $m \geq 2$, we define the $m$-th shell as
\begin{align}
\text{Shell}(m) = \{ v \in \Lambda_{24} \quad \mid \quad \|v\|_2^2 = 2m \}.
\end{align}
The minimal squared norm of $\Lambda_{24}$ is $4$ (for $m=2$). The disjoint shells partition the full lattice:
{\setlength{\abovedisplayskip}{2pt}
\setlength{\belowdisplayskip}{1pt}
\setlength{\abovedisplayshortskip}{0pt}
\setlength{\belowdisplayshortskip}{1pt} 
\setlength{\jot}{3pt} 
\begin{align}
\Lambda_{24} = \bigcup_{m=2}^{\infty} \text{Shell}(m), \quad \text{and define ball-cuts }
\Lambda_{24}(M) = \bigcup_{m=2}^{M} \text{Shell}(m).
\end{align}
up to squared norm $M=2m$. Since shells are disjoint, the cardinality is a cumulative sum:
\begin{align}
\begin{split}
N(M) = |\Lambda_{24}(M)| = \sum_{m=2}^{M} n(m),\quad \text{with }
n(m) = |\text{Shell}(m)|
\end{split}
\end{align}
}
\begin{wraptable}{r}{0.6\linewidth}
\centering
\vspace{-1.5em}
\caption{Shell structure of the Leech lattice $\Lambda_{24}$.}
\vspace{0.5em}
\resizebox{\linewidth}{!}{
\begin{tabular}{c | c r r c }
\textbf{$\,m\,$} & \textbf{Radius $\sqrt{2m}$} & \textbf{Shell cardinality} $n(m)$ & \textbf{Cumulative count} $N(m)$ & \textbf{Bits / dim} \\
\hline
2 & 2 & 196{,}560 & 196{,}560 & 0.75 \\
3 & $\sqrt{6}$ & 16{,}773{,}120 & 16{,}969{,}680 & 1.042 \\ 
4 & 2$\sqrt{2}$ & 398{,}034{,}000 & 415{,}003{,}680 & 1.208 \\
5 & $\sqrt{10}$ & 4{,}629{,}381{,}120 & 5{,}044{,}384{,}800 & 1.375 \\
. & & & \\
\textbf{13} & $\sqrt{26} $ & 16{,}993{,}109{,}532{,}672 & 280{,}974{,}212{,}784{,}720 & \textbf{2.000} \\
. & & & \\
19 & $\sqrt{38} $ & 1{,}104{,}550{,}081{,}689{,}600 & 23{,}546{,}209{,}100{,}646{,}960 & 2.292 \\
\end{tabular}
}
\label{tab:shell-sizes}
\vspace{-1.0em}
\end{wraptable}
These shell sizes are the coefficients of the Leech lattice theta series \citep{leech1967notes}, but for our purposes they can be treated as fixed combinatorial quantities that can be precomputed or tabulated. Since $M$ controls the size of the bounded lattice subset, it can be thought of as a parameter that governs both the bitrate and the effective fineness of the quantization grid. Table~\ref{tab:shell-sizes} reports the number of points per shell $n(m)$, cumulative number of points $N(M)$ together with the required bits per dimension $\lceil \log_2 N(M) \rceil / 24$ needed to store an index identifying lattice points within the bounded set.
An alternative way to construct a finite subset from a lattice, beyond spherical shaping, is the \textit{shape–gain} approach \citep{sabin1982product, hamkins2002gaussian}, in which vector magnitudes and directions are quantized separately. The magnitude is handled by a standard scalar quantizer, while the direction is mapped to a \textit{spherical code}, i.e., a finite set of points on the high‑dimensional unit sphere. Such spherical codes can be constructed by normalizing all lattice points in one or more shells. For shape–gain, it is not so much the packing density of a lattice, but rather the uniformity of the constructed spherical code that determines the quantization performance. Interestingly, the Leech lattice excels on both fronts: it has exceptionally high packing density \citep{conway2013sphere}, and its shells yield remarkably uniform spherical codes. In \autoref{app:spherical-bounding-vs-shape-gain}, we compare spherical shaping with shape–gain quantization for Gaussian sources using the Leech lattice and find that both perform very well, with shape–gain achieving slightly improved rate–distortion in our setting, motivating its use.

One may also wonder whether it is preferable to form a spherical code using individual shells or using a cumulative union of shells. We investigate this in \autoref{app:union-of-shells} and find that using the union of shells up to $m$ yields a more uniform spherical code, by a lower empirical angular error to the closest point per bitwidth, compared to using individual shells. We conjecture that this trend persists, and that cumulative unions of shells achieve strictly lower distortion than the rate–distortion at infinite rate limits. This motivated our use of cumulative shell unions in shape-gain.

\subsection{Extended Golay Code construction of $\Lambda_{24}$}

\begin{wraptable}{r}{0.45\linewidth}
\vspace{-4.5em}
\caption{Composition of lattice vectors grouped by equivalence class. For each class, the columns indicate the multiplicities of coordinates equal to \(\pm 7, \pm 6, \dots, \pm 0\), describing the canonical coordinate pattern of vectors in that class. Shell cardinalities increase with \(m\).}
\centering
\setlength{\tabcolsep}{3pt}
\renewcommand{\arraystretch}{0.9}
\vspace{1em}
\resizebox{\linewidth}{!}{
\begin{tabular}{c c r | c c c c c c c c c}
m & parity & count & $\pm8$ & $\pm7$ & $\pm$6 & $\pm$5 & $\pm$4 & $\pm$3 & $\pm$2 & $\pm$1 & $\pm$0 \\
\hline
2 & even & \small{1104} & 
\textcolor{mygray}{0} &
\textcolor{mygray}{0} &
\textcolor{mygray}{0} &
\textcolor{mygray}{0} & 2 & \textcolor{mygray}{0} & \textcolor{mygray}{0} & \textcolor{mygray}{0} & 22 \\
 & even & \small{97152} & 
\textcolor{mygray}{0} &
\textcolor{mygray}{0} &
 \textcolor{mygray}{0} & \textcolor{mygray}{0} & \textcolor{mygray}{0} & \textcolor{mygray}{0} & 8 & \textcolor{mygray}{0} & 16 \\
\cdashline{4-12}[5pt/4pt]
 & odd & \small{98304} & 
\textcolor{mygray}{0} &
\textcolor{mygray}{0} &
 \textcolor{mygray}{0} & \textcolor{mygray}{0} & \textcolor{mygray}{0} & 1 & \textcolor{mygray}{0} & 23 & \textcolor{mygray}{0} \\
\hline
3 & even & \small{3108864} & 
\textcolor{mygray}{0} &
\textcolor{mygray}{0} &
\textcolor{mygray}{0} & \textcolor{mygray}{0} & 1 & \textcolor{mygray}{0} & 8 & \textcolor{mygray}{0} & 15 \\
 & even & \small{5275648} & 
\textcolor{mygray}{0} &
\textcolor{mygray}{0} &
 \textcolor{mygray}{0} & \textcolor{mygray}{0} & \textcolor{mygray}{0} & \textcolor{mygray}{0} & 12 & \textcolor{mygray}{0} & 12 \\
\cdashline{4-12}[5pt/4pt]
 & odd & \small{98304} & 
\textcolor{mygray}{0} &
\textcolor{mygray}{0} &
 \textcolor{mygray}{0} & 1 & \textcolor{mygray}{0} & \textcolor{mygray}{0} & \textcolor{mygray}{0} & 23 & \textcolor{mygray}{0} \\
 & odd & \small{8290304} & 
\textcolor{mygray}{0} &
\textcolor{mygray}{0} &
 \textcolor{mygray}{0} & \textcolor{mygray}{0} & \textcolor{mygray}{0} & 3 & \textcolor{mygray}{0} & 21 & \textcolor{mygray}{0} \\
\hline
4 & even & \small{170016} & 
\textcolor{mygray}{0} &
\textcolor{mygray}{0} &
\textcolor{mygray}{0} & \textcolor{mygray}{0} & 4 & \textcolor{mygray}{0} & \textcolor{mygray}{0} & \textcolor{mygray}{0} & 20 \\
 & even & \small{48} & 
1 &
\textcolor{mygray}{0} &
 \textcolor{mygray}{0} & \textcolor{mygray}{0} & \textcolor{mygray}{0} & \textcolor{mygray}{0} & \textcolor{mygray}{0} & \textcolor{mygray}{0} & 23 \\
 & even & \small{46632960} & 
\textcolor{mygray}{0} &
\textcolor{mygray}{0} &
 \textcolor{mygray}{0} & \textcolor{mygray}{0} & 2 & \textcolor{mygray}{0} & 8 & \textcolor{mygray}{0} & 14 \\
 & even & \small{777216} &
\textcolor{mygray}{0} &
\textcolor{mygray}{0} & 1 & 
 \textcolor{mygray}{0} & \textcolor{mygray}{0} & \textcolor{mygray}{0} & 7 & \textcolor{mygray}{0} & 16 \\
 & even & \small{126615552} & 
\textcolor{mygray}{0} &
\textcolor{mygray}{0} &
 \textcolor{mygray}{0} & \textcolor{mygray}{0} & 1 & \textcolor{mygray}{0} & 12 & \textcolor{mygray}{0} & 11 \\
 & even & \small{24870912} & 
\textcolor{mygray}{0} &
\textcolor{mygray}{0} &
 \textcolor{mygray}{0} & \textcolor{mygray}{0} & \textcolor{mygray}{0} & \textcolor{mygray}{0} & 16 & \textcolor{mygray}{0} & 8 \\
\cdashline{4-12}[5pt/4pt]
 & odd & \small{24870912} & 
\textcolor{mygray}{0} &
\textcolor{mygray}{0} &
 \textcolor{mygray}{0} & 1 & \textcolor{mygray}{0} & 2 & \textcolor{mygray}{0} & 21 & \textcolor{mygray}{0} \\
 & odd & \small{174096384} & 
\textcolor{mygray}{0} &
\textcolor{mygray}{0} &
 \textcolor{mygray}{0} & \textcolor{mygray}{0} & \textcolor{mygray}{0} & 5 & \textcolor{mygray}{0} & 19 & \textcolor{mygray}{0} \\
\end{tabular}
}
\vspace{-2em}
\label{tab:leaders-table}
\end{wraptable}

For efficient search in $\Lambda_{24}$, we follow \citep{adoul1988nearest} and
use the classical construction of the Leech lattice from the extended Golay
code \citep{conway2013sphere}. This representation organizes the lattice as an
implicit hierarchy of integer vectors. The extended Golay code
$\mathcal{G}_{24} \subset \mathbb{F}_2^{24}$ is the unique binary code of size $4096$, whose nonzero codewords have Hamming weights
in $\{8,12,16,24\}$. 

Vectors in $\Lambda_{24}$ are first grouped by \emph{shell}, and within shells, vectors share the same unordered multiset of absolute values form a \emph{class}, represented by a canonical leader up to admissible permutations and sign choices. Each class is intrinsically \emph{even} or \emph{odd} (whether it is a subset of $L^{\text{even}}$ or $L^{\text{odd}}$, defined shortly), determining the permissible coordinate permutations and sign assignments, dictated by the underlying Golay structure. Our proposed indexing scheme follows the same hierarchy. For example, the first shell contains $196{,}560$ lattice points, of which the first $1{,}104$ belong to the first class, consisting of vectors with two coordinates equal to 4 and twenty-two equal to 0 (see \autoref{tab:leaders-table}). The (nearest neighbour) quantization- and dequantization algorithms exploit the hierarchy to avoid explicit enumeration of lattice points.

\paragraph{Integer-coordinate formulation.}
The Leech lattice can be defined as the union of:
{\setlength{\abovedisplayskip}{0pt}
\setlength{\belowdisplayskip}{0pt}
\setlength{\abovedisplayshortskip}{0pt}
\setlength{\belowdisplayshortskip}{1pt} 
\setlength{\jot}{3pt} 
\begin{align}
L^{\mathrm{int}} = L^{\mathrm{even}} \,\cup\, L^{\mathrm{odd}},
\qquad
\Lambda_{24} = \frac{1}{\sqrt{8}}\, L^{\mathrm{int}} \subset \mathbb{R}^{24},
\end{align}
}
a scaled union of two integer-valued sets which we call the \textit{even} and \textit{odd} cosets,
\begin{align}
\begin{split}
L^{\mathrm{even}}
&{=}\left\{
x \in \mathbb{Z}^{24}
\;\middle|\;
{\tiny
\begin{aligned}
&\text{(i) } x_i \equiv 0 \pmod{2} \\
&\text{(ii) } (x/2) \bmod 2 \in \mathcal{G}_{24} \\
&\text{(iii) } \sum_i x_i \equiv 0 \pmod{8}
\end{aligned}
}
\right\}
\end{split}
\begin{split}
L^{\mathrm{odd}}
&{=}\left\{
x \in \mathbb{Z}^{24}
\;\middle|\;
{\tiny
\begin{aligned}
&\text{(i) } x_i \equiv 1 \pmod{2} \\
&\text{(ii) } \bigl((x-\mathbf{1})/2\bigr) \bmod 2 \in \mathcal{G}_{24} \\
&\text{(iii) } \sum_i x_i \equiv 4 \pmod{8}
\end{aligned}
}
\right\}
\end{split}
\end{align}
where the roles of the three constraints are:
\begin{itemize}
\vspace{-0.8em}
    \item[(i)] \emph{Parity constraint:} forces vector in even or odd coset.
\vspace{-0.5em}
    \item[(ii)] \emph{Golay constraint:} ensures mod-$2$ reductions of halved vectors match codewords of $\mathcal{G}_{24}$.
\vspace{-0.5em}
    \item[(iii)] \emph{Sum constraint:} enforces the global congruence needed for the resulting lattice to be even.
\vspace{-0.8em}
\end{itemize}
and we write $\mathbf{1}=(1,\dots,1)\in\mathbb{Z}^{24}$. With the $1/\sqrt{8}$ normalization, the lattice is even and unimodular. 

%
\subsection{Class Structure and Leaders}
\label{sec:leaders}

Each shell $\mathrm{Shell}(m)$ can be decomposed into different \emph{classes}, where we define a class as a set of all lattice points obtainable from one another by the allowed permutations of coordinates and sign flips. \autoref{tab:leaders-table} lists the classes in the first three shells of the Leech lattice, together with their cardinalities.

For convenience, we select a single representative from each equivalence class under coordinate permutations; typically this is the point whose coordinates have been reordered so that their absolute values appear in a fixed canonical order. We refer to this representative as the \emph{leader}. In the integer embedding, these leaders admit a compact combinatorial description: up to permutation of coordinates, each class is characterized by a multiset
{\setlength{\abovedisplayskip}{2pt}
\setlength{\belowdisplayskip}{1pt}
\setlength{\abovedisplayshortskip}{0pt}
\setlength{\belowdisplayshortskip}{1pt} 
\setlength{\jot}{3pt} 
\begin{align}
\{\, a_1^{p_1},\, a_2^{p_2},\,\dots,\, a_k^{p_k} \,\}, \qquad \sum_i p_i=24,
\quad \text{ where } a_i \in \tfrac{1}{\sqrt{8}}\mathbb{Z}
\vspace{-1em}
\end{align}
}
Absolute coordinates $a_i$ and exponents $p_i$ fully specify vector coordinate compositions of each class.

\paragraph{Even and Odd Classes}
\label{sec:even-and-odd-leaders}

Each class lies entirely in either $L^{\mathrm{even}}$ or $L^{\mathrm{odd}}$, and this determines how its nonzero coordinates may be placed and signed. A leader fixes the multiset of absolute coordinate values and whether the class is even or odd. What remains unspecified is how those fixed coordinate values are placed across the \(24\) positions and which sign patterns are allowed. Given a Golay codeword \(c \in \mathcal{G}_{24}\), define
{\setlength{\abovedisplayskip}{4pt}
\setlength{\belowdisplayskip}{4pt}
\setlength{\abovedisplayshortskip}{0pt}
\setlength{\belowdisplayshortskip}{1pt} 
\setlength{\jot}{3pt} 
\begin{align}
F_0(c) {=} \{\, i : c_i {=} 0 \,\},
F_1(c) {=} \{\, i : c_i {=} 1 \,\}.
\end{align}
}
For \textbf{even classes}, admissible placements arise only from Golay codewords \(c\)
whose Hamming weight matches the number of coordinates of the leader
congruent to \(2 \pmod{4}\). For any such \(c\), coordinates with
\(x_i \equiv 0 \pmod{4}\) must lie in \(F_0(c)\), and those with
\(x_i \equiv 2 \pmod{4}\) must lie in \(F_1(c)\), up to permutation. Signs of
coordinates in \(F_0(c)\) are unrestricted because the \(0 \pmod{4}\)
congruence is invariant under sign flips. Signs in \(F_1(c)\) are constrained
only by the global condition \(\sum_i x_i \equiv 0 \pmod{8}\), which fixes
the parity of the number of negative signs among the \(F_1(c)\) entries.
Thus, even classes admit only those placements induced by compatible Golay
codewords and a correspondingly restricted set of sign patterns. For \textbf{odd classes}, \emph{every} codeword \(c \in \mathcal{G}_{24}\) yields a valid placement.
The congruence conditions then determine the coordinate types uniquely:
positions in \(F_0(c)\) carry entries with \(x_i \equiv 1 \pmod{4}\),
and positions in \(F_1(c)\) carry entries with \(x_i \equiv 3 \pmod{4}\).
Consequently, the sign pattern is fixed by these congruences (up to an overall
sign flip). 


\subsection{Subclass Counts and Class Cardinality}
\label{sec:subclasses}
The cardinality of each class follows directly from combinatorial coordinate placements, signs, and coordinate multiplicities. Fix a leader with condensed multiset \(\{a_1^{p_1},\dots,a_k^{p_k}\}\). The number of lattice points in its class factorizes into: (a) the number \(A\) of
Golay codewords \(c \in \mathcal{G}_{24}\) that yield an admissible placement for
this leader (odd classes have \(A=4096\); even classes have \(A\in\{1,759,2576,759,1\}\)
depending on the required weight), (b) the number \(2^{B}\) of admissible sign
assignments consistent with parity and the global mod--\(8\) sum constraint, (c)
the multinomial factor \(\frac{24!}{\prod_i p_i!}\) accounting for permutations of
coordinates with identical absolute values, and (d) an additional divisor
\(\prod_j q_j!\) capturing permutations that act trivially because equal-valued
coordinates fall within the same Golay subset \(F_0(c)\) or \(F_1(c)\).
The class cardinality is $ A \cdot 2^{B}
\cdot
\frac{24!}{\prod_{i=1}^k p_i!}
\cdot
\frac{1}{\prod_j q_j!}.
$.

\vspace{-0.5em}
\section{Leech Lattice Vector Quantization (LLVQ)}
\label{sec:method}

We build upon the fast exact nearest neighbour search for single Leech lattice shells by \citet{adoul1988nearest}. This method generates candidates via leaders (the canonical absolute-value patterns) and uses Golay-derived placements, parity-constrained sign patterns, to rank dot products with the input. On a single shell, this dot-product ranking coincides with Euclidean ranking (see Sec.~\ref{sec:spherical-search} below). We generalize and extend the algorithm in two important ways: (i) we enable nearest neighbour search over multiple $\Lambda_{24}(m)$ shells, where candidate norms vary and Euclidean vs.\ angular scoring no longer coincide; and allow support for both \emph{Euclidean} scoring (for spherical shaping) and \emph{angular} scoring (for shape--gain), discussed in \S\ref{sec:spherical-search}; and (ii) we introduce a \emph{bijective indexing mechanism} aligned with the Leech lattice hierarchy (shells, classes, and local symmetries), yielding compact integer codes and exact reconstruction through the dequantizer (see \S\ref{sec:indexing-scheme}, \S\ref{sec:dequantizer}). Finally, we describe \emph{Spherical GPTQ}, a norm-preserving GPTQ variant for directional codes like LLVQ (\S\ref{sec:spherical-gptq-main}).

\subsection{Extending to Spherical Search in $\Lambda_{24}(m)$}
\label{sec:spherical-search}

In the original single-shell setting of \citet{adoul1988nearest}, all candidates share the same norm so Euclidean and angular distances induce the same ordering:
$
\|x {-} v\|^2 {=} \|x\|^2 {+} \|v\|^2 {-} 2\langle x, v\rangle \text{\ (fixed $\|v\|$)}
$
Consequently, ranking by $-\langle x,v\rangle$ is equivalent to minimizing $\|x{-}v\|$. For LLVQ we generalize the search to multiple shells of $\Lambda_{24}(m)$. Since candidate norms vary across shells the Euclidean vs.\ angular equivalence no longer holds. We add support for two metrics: \emph{Euclidean distance} (for spherical shaping), and \emph{angular distance} via cosine similarity (for shape--gain). This can be implemented by normalizing both the input and candidates, $\hat{x}{=}x/\|x\|$ and $\hat{v}{=}v/\|v\|$ and maximizing $\langle \hat{x}, \hat{v}\rangle$.

\subsection{Indexing Scheme}
\label{sec:indexing-scheme}

While nearest neighbour algorithm by \citet{adoul1988nearest} introduced an elegant exact search algorithm over leaders, placements, and signs, it does not provide a bijective indexing scheme, which is required for quantization-based compression. For compression, each lattice vector must correspond to a unique integer index (or bitstring), and this mapping must be efficiently invertible. We therefore extend the procedure to allow indexing of Leech lattice vectors aligned with its described hierarchical structure. First, shells are ordered by increasing radius: the first 196,560 indices correspond to the first shell, the next 16,773,120 indices correspond to the second shell, and so on. Within each shell, we assign consecutive index ranges to the classes (e.g., based on cardinality or any other fixed, consistent ordering). Inside each class, the remaining degrees of freedom: permutations, sign flips, or other symmetries, are indexed locally. The local class index is combined with the shell and class indices through standard index linearization (as used when flattening a multidimensional array into a 1‑dimensional array). This yields a unique and invertible global index for every vector. The inverse mapping follows the usual unflattening procedure: integer division recovers the shell and class, and modulo recovers the index inside the class.

We index vectors according to the natural hierarchy of shells, classes, and
intra-class ordering.
\vspace{-0.4em}
\paragraph{(1) Shell level}
Shells are ordered by increasing squared norm. Let $n(m)=|\mathrm{Shell}(m)|$
and $N(m)=\sum_{\ell \le m} n(\ell)$ be cumulative offsets. The global indices
$\{0,\dots,N(2)-1\}$ enumerate $\mathrm{Shell}(2)$, the next
$\{N(2),\dots,N(3)-1\}$ enumerate $\mathrm{Shell}(3)$, and so on.
\vspace{-0.4em}
\paragraph{(2) Class level}
Within shell $m$, we fix a deterministic total order over its classes (e.g.,
lexicographic on leaders, then parity, then a fixed tie-break) and assign to
each class a contiguous subrange whose length equals its cardinality. Per shell, leaders, class sizes, and cumulative offsets are stored solely to support dequantization (shell/class lookup and vector reconstruction from indices).
\vspace{-0.4em}
\paragraph{(3) Local symmetry level}
Within each class, the local index is obtained by decomposing the integer into
successive choices: first the Golay refinement, then the sign pattern, and
finally the permutation coset. This is implemented by repeated modulo and
integer-division operations.

\subsection{Spherical GPTQ: A geometric interpretation of Scale-corrected Shape-Gain with GPTQ}
\label{sec:spherical-gptq-main}

When LLVQ is used in a shape--gain regime, the quantizer primarily determines a vector direction, while magnitudes are handled by an explicit gain quantizer. In practice, shape--gain pipelines are often coupled with a scale-correction step (App.~\ref{sec:qd-blocks}): after quantizing the direction, the reconstruction is rescaled to minimize Euclidean distance with the original vector.

We observe that, when this scale correction is interleaved with GPTQ-style Hessian error feedback, it admits a clean geometric interpretation: the gain reset is precisely a \emph{retraction} onto a product of constant-norm manifolds (hypersphere). This yields a \emph{Spherical GPTQ} viewpoint in which the greedy PTQ dynamics operate predominantly on directions. Concretely, given a block $\vw$ and a candidate $\tilde{\vw}=\mathcal{Q}(\vw)$, the gain reset can be written as
\begin{equation}
\hat{\vw}
\;=\;
\Pi_{\|\vw\|}(\tilde{\vw})
\;:=\;
\frac{\|\vw\|_2}{\|\tilde{\vw}\|_2}\,\tilde{\vw},
\label{eq:spherical_retraction}
\end{equation}
and GPTQ residuals are formed using $\hat{\vw}$ rather than $\tilde{\vw}$. We refer to this combined procedure as \emph{Spherical GPTQ}. The interpretation also enables a clean ablation showing that the Leech codebook and norm-preserving corrections each improve PTQ in isolation and are complementary when combined (further details and ablations in \autoref{sec:spherical-gptq}).

\section{SQNR and Retention on Gaussian Source}

\begin{wrapfigure}{r}{0.35\linewidth}
\vspace{-5em}
    \centering
    \resizebox{\linewidth}{!}{
    \includegraphics{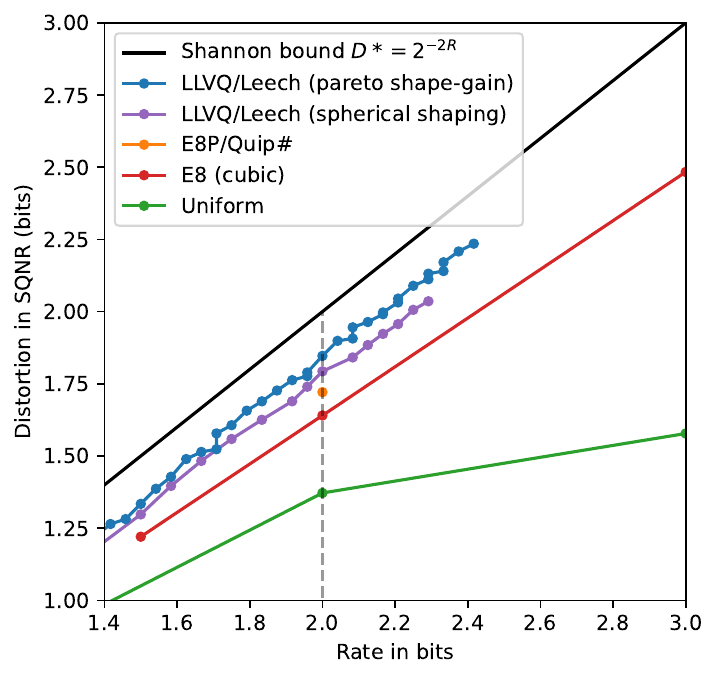}
    }
\vspace{-2em}    \caption{$\widehat{\text{SQNR}}_{\text{bits}}$ versus bitrate (bits/dim) on Gaussian source.}
\vspace{-1.0em}
    \label{fig:rate_distortion}
\end{wrapfigure}

We evaluate quantization performance by generating i.i.d.\ samples $w \sim \mathcal{N}(0, 1)$ from a zero-mean, unit-variance Gaussian distribution, applying the quantization scheme at different bitrates, and measuring the resulting distortion. The distortion is computed as the mean squared error (MSE) between the original and quantized samples. For $n$ samples $\{w_i\}_{i=1}^n$ and quantizer $q(\cdot)$, we can obtain an unbiased estimate of the empirical quantization error per weight {\setlength{\abovedisplayskip}{2pt}
\setlength{\belowdisplayskip}{1pt}
\setlength{\abovedisplayshortskip}{0pt}
\setlength{\belowdisplayshortskip}{1pt} 
\setlength{\jot}{3pt} 
\begin{align}
\widehat{\text{MSE}} \;=\; \frac{1}{n}\sum_{i=1}^n \| w_i - q(w_i) \|^2_2 / D,
D = \text{dim}(w_i)
\end{align}
and the empirical 
$
\widehat{\text{SQNR}}_{\text{bits}}
\;=\; -\frac{1}{2}\,\log_2\!\big(\widehat{\text{MSE}}\big).
$
For an ideal Gaussian source, the Shannon rate--distortion function gives the minimum achievable MSE at rate $R$ as
$
\text{MSE}^*(R) \;=\; \sigma^2\,2^{-2R}.
$
For our unit-variance source ($\sigma^2 = 1$), this becomes $\text{MSE}^*(R)=2^{-2R}$. Using the same convention for SQNR in bits, the optimal (Shannon) SQNR is
$
\text{SQNR}^*_{\text{bits}}(R)
\;=\; -\frac{1}{2}\,\log_2\!\big(2^{-2R}\big)
\;=\; R.
$
Thus, following this convention, the Shannon limit for a Gaussian source corresponds to the straight line $\,\text{SQNR}^*_{\text{bits}}(R)=R\,$ (i.e., a $y = x$ line) in the SQNR--versus--rate plot. To convert to the other common base-10 decibel (dB) unit, multiply $\text{SQNR}_{\text{bits}}$ with $20\log_{10}(2)\!\approx\!6.0206$.

Due to structural constraints, practical quantizers inevitably fall below the theoretical limit. \autoref{fig:rate_distortion} shows SQNR versus bitrate behavior on an idealistic Gaussian source. Among the evaluated methods, our LLVQ construction attains the highest empirical SQNR at each tested bitrate, consistently outperforming existing approaches and tracking the Shannon bound more closely than the baselines.

We quantify closeness to the Shannon limit by measuring \emph{retention}:
$\text{Ret} (\%) = \frac{\widehat{\text{SQNR}}_{\text{bits}}}{\text{SQNR}^*_{\text{bits}}(R)} {\times} 100
$
with conveniently $\text{SQNR}^*_{\text{bits}}(R)=R$, under our convention of measuring SQNR in bits.

\begin{wraptable}{l}{0.6\linewidth}
\vspace{-2em}
\caption{Gaussian information retention at 2 bits/dim.}
\vspace{0.5em}
\label{tab:gaussian_performance}
    \centering
\resizebox{0.8\linewidth}{!}{
    \begin{tabular}{l | c l c | c c}
        \textbf{Method}
        & \textbf{Dim}
        & \textbf{$\widehat{\text{MSE}}$} $\downarrow$
        & \textbf{$\widehat{\text{SQNR}}$} (bits) $\uparrow$
        & \textbf{Ret} (\%) $\uparrow$ \\
        \hline
Uniform
& 1 
& 0.15
& 1.37 & 69 \\
Lloyd-Max
& 1 
& 0.12
& 1.53 & 77 \\
\cdashline{0-5}[5pt/4pt] 
E8 (cubic) 
& 8
& 0.103
& 1.64 & 82.0 \\
LLVQ/Leech [spherical shaping] (\textbf{Ours})
& 24
& \textbf{0.084}
& \textbf{1.79} & \textbf{89.4} \\
LLVQ/Leech [shape-gain] (\textbf{Ours})
& 24
& \textbf{0.078}
& \textbf{1.84} & \textbf{92.1} \\
\hline
Theoretical limit & - & 0.0625 & 2 & 100 \\
\hline
    \end{tabular}
}
\end{wraptable}

As summarized in Table~\ref{tab:gaussian_performance}, the empirical SQNR and retention scores at $R=2$~bits/dim reveal clear gaps in performance across quantizers. Uniform quantization performs worst, reflecting its known suboptimality in Gaussian sources, while structured lattice-based schemes such as cubically bounded E8 and E8P/Quip \# achieve notably higher retention. Our LLVQ constructions achieve the best performance, achieving MSE and SQNR closest to the Shannon limit. This shows that LLVQ uses the available bitrate more efficiently, yielding substantially lower distortion on Gaussian inputs than existing lattice-based approaches.

\section{LLM Quantization Results}


For empirical results, we compute (GPTQ-style) layer-wise Hessians on 6,100 sequences from DCLM-edu \citep{li2024datacomp,allal2025smollm2}, matching the calibration set size used in prior work \citep{tseng2024quip}. When finetuning is applied, we update only the input scales, scalars shared across rows, which add a negligible number of bits per weight and relatively little risk of overfitting.
\subsection{PTQ Results (Own Pipeline)}
\begin{table*}[h]
\vspace{-2em}
\centering
\caption{Comparison with wider set of methods for quantizing Llama-v2 7B model using different quantization methods, evaluated by wikitext-2 (Wiki) perplexity at 4096 context length and downstream task performance (CSR and MMLU), comparing results in literature. LLVQ consistently outperforms standard vector quantization approaches in performance against bits per parameter (Bits).}
\label{tab:ptq-performance-literature}
\resizebox{\linewidth}{!}{
\begin{tabular}{l l r c c | c c c c c c c }
& & & & & \multicolumn{7}{c}{Llama-2 7B} \\
\textbf{Method} & \textbf{Details} (from literature) & \textbf{Finetuned} & \textbf{LM Eval} & \,\textbf{BPW}\, &
\textbf{Wiki} $\downarrow$ 
& \textbf{Arc-C} $\uparrow$  & \textbf{Arc-E} $\uparrow$ & \textbf{BoolQ} $\uparrow$ & \textbf{Winogrande} $\uparrow$ & \textbf{Hellaswag} $\uparrow$ & \textbf{PiQA} $\uparrow$ \\
\hline
Baseline & (\textbf{Ours}) & - & acc & 16 & 5.11
& \textbf{43.2} & \textbf{75.6} & \textbf{79.3} & \textbf{69.9} & \textbf{57.1} & \textbf{78.1}  \\
\hline
Quip\# & Tables 4 \& 10 of Quip\# paper & No & acc/acc\_norm & 2 & 8.22 & 32.5 & 42.8 & 62.3 & 62.4 & - & 71.2 \\ 
\hline
LLVQ (\textbf{Ours}) &
[spherical shaping]
& No & acc & 2 & 7.61 & 34.7 & 69.3 & 67.5 & 64.6 & 46.6 & 73.5 \\
LLVQ (\textbf{Ours}) &
[shape-gain, 2 bit gain]
& No & acc & 2 & \textbf{6.83} & 
\textbf{35.5} & \textbf{69.8} & \textbf{73.0} & \textbf{66.9} & \textbf{49.7} & \textbf{75.2} \\
\hline
\hline
Quip &  Table 3 of Quip\# paper & Yes & acc & 2 & - & 19.4 & 26.0 & 54.6 & 51.8 & - & - \\
OmniQ & Table 3 of Quip\# paper & Yes & acc & 2 & - & 21.6 & 35.2 & 57.5 & 51.5 & - & - \\
AQLM & Tables 5 \& 6 of QTIP paper & Yes & not reported & 2.07 & 6.93 & 32.8 & 63.7 & 74.8 & 65.7 & - & - \\
Quip\# & Tables 5 \& 6 of QTIP paper & Yes & not reported & 2 & 6.19 & 35.2 & 65.3 & 75.4 & 64.9 & - & - \\
QTIP & Tables 5 \& 6 of QTIP paper & Yes & not reported & 2 & 5.86 & 35.7 & 65.6 & \textbf{75.9} & 64.7 & - & - \\
PV-tuning & Table 8 of PV-tuning paper & Yes & acc & 2 & 5.84 & 38.4 & 71.2 & - & \textbf{66.7} & 53.5 & 77.0 \\
\hline
LLVQ (\textbf{Ours}) &
[spherical shaping]
& Yes & acc & 2 & \textbf{5.60} & \textbf{40.6} & \textbf{72.9} & 70.9 & 65.1 & 52.5 & 75.5 \\
LLVQ (\textbf{Ours}) &
[shape-gain, 2 bit gain]
& Yes & acc & 2 & \textbf{5.48} & \textbf{39.8} & \textbf{72.9} & 75.3 & 66.3 & \textbf{54.1} & \textbf{77.1} \\
\hline
\end{tabular}
\vspace{-6em}
}
\end{table*}

We begin by evaluating post‑training quantization (PTQ) using a unified quantization pipeline enabling a strict apples‑to‑apples comparison across methods. Table~\ref{tab:other-models} summarizes PTQ results on Llama‑2 7B, Llama-3 8B, Ministral-3 8B, and Qwen‑v3 4B and 8B under this setup. The gap between scalar baselines (RTN, GPTQ \citep{frantar2022gptq}, Quarot \citep{ashkboos2024quarot}) and higher‑dimensional VQ methods is immediately evident: naive 2‑bit RTN yields extremely high perplexity and severe task degradation, while GPTQ and Quarot improve stability through Hessian curvature and Hadamard rotations, yet remain constrained by the limited representational capacity of 1D quantization.

Higher‑dimensional approaches, such as Quip\#, substantially reduce this gap. However, our LLVQ method, employing Leech‑lattice‑based 24‑dimensional vector quantization, consistently achieves the strongest 2‑bit performance across all metrics. Wikitext perplexity, MMLU accuracy, and CSR all show clear gains over Quip\#, demonstrating that high‑dimensional structured lattices provide markedly more efficient weight‑space packing than both scalar and E8‑based quantizers.

\subsection{Spherical GPTQ and Hadamard-Free PTQ}

Following prior work \citep{ashkboos2024quarot, chee2023quip}, we consider randomized Hadamard rotations on layer inputs/outputs, which often improve quantization but may require online application when not fusible into weights, increasing latency. We additionally evaluate \emph{Spherical GPTQ}, a norm-preserving GPTQ variant obtained by applying the standard shape--gain scale-corrections within the GPTQ loop (see Sec.~\ref{sec:spherical-gptq}). Empirically, we find Spherical GPTQ substantially reduces the reliance on Hadamard/rotation preprocessing, enabling strong Hadamard-free (or Hadamard-reduced) PTQ for LLVQ (Table~\ref{tab:hadamard-ablation}).

\begin{wraptable}{r}{0.5\linewidth}
\vspace{-4em}
\centering
\caption{Sherical GPTQ interpretation and Hadamard-free PTQ: ablation with and without Hadamard rotations, evaluated on Wikitext-2 perplexity (PPL) and downstream tasks (CSR and MMLU). LLVQ remains competitive without rotations when norm-preserving corrections are used, indicating reduced reliance on online transforms.}
\label{tab:hadamard-ablation}
\resizebox{\linewidth}{!}{
\begin{tabular}{l | c c l | c c c }
& & & & \multicolumn{3}{c}{Llama-2 7B} \\
\textbf{Method} \small{(no finetune)} & \textbf{Dim}&
\footnotesize{\textbf{BPW}} & 
\textbf{Hadamard}
& \textbf{Wiki} $\downarrow$ & \textbf{MMLU} $\uparrow$ & \textbf{CSR} $\uparrow$ \\
\hline
Baseline & 1 
& 16 & - & 5.12 & 45.7 & 70.4 \\
\hline
Integer (GPTQ) & 1 
& 2
& No Rotation & 
3411.6 & 26.6 & 39.7 \\
Integer (Quarot) & 1 & 
2 & 
Input & 41.87 & 27.0 & 41.7 \\
Integer & 1 
&
2 &
\small{Input + Output} &
37.83 & 26.1 & 48.4
\\
\cdashline{0-6}[5pt/4pt] 
E8P & 8 &
2 &
No Rotation & 105.98 & 24.8 & 44.9 \\
E8P & 8 & 
2 &
Input & 9.24 & 31.0 & 59.8 \\
E8P (Quip\#) & 8 &
2 &
\small{Input + Output} & 7.96 & 30.5 & 61.4
\\
\cdashline{0-6}[5pt/4pt] 
\cdashline{0-6}[5pt/4pt] 
LLVQ [spherical shaping] 
& 24 & 2 &
No Rotation & 191.90 & 24.0 & 53.5 \\
LLVQ [spherical shaping] 
& 24 & 2 & 
Input & \textbf{6.80} & \textbf{35.1} & \textbf{65.4}
 \\
LLVQ [spherical shaping] 
& 24 & 2 &
\small{Input + Output} & \textbf{7.61} & \textbf{33.4} & \textbf{62.1} \\
\cdashline{0-6}[5pt/4pt] 
LLVQ [shape-gain, 2 bit gain] 
& 24 & 2 & 
No Rotation & \textbf{7.27} & \textbf{29.8} & \textbf{61.5} \\
LLVQ [shape-gain, 2 bit gain] 
& 24 & 2 &
Input & 
\textbf{6.90} & \textbf{36.0} & \textbf{63.6}
\\
LLVQ [shape-gain, 2 bit gain] 
& 24 & 2 &
\small{Input + Output} & \textbf{6.83} & \textbf{34.9} & \textbf{64.6}
 \\
\hline
\end{tabular}
}
\vspace{-1.0em}
\end{wraptable}

\paragraph{Hadamard-free PTQ.}
Table~\ref{tab:hadamard-ablation} summarizes an ablation over (i) scalar integer baselines, (ii) E8P/Quip\#, and (iii) LLVQ, under different preprocessing choices. Two trends emerge. First, rotations strongly benefit scalar quantization and low-dimensional codes, consistent with their limited representational flexibility. Second, LLVQ remains competitive \emph{without} rotations when paired with the Spherical GPTQ interpretation: norm preservation eliminates the dominant failure mode (radial drift), making Hadamard preprocessing far less critical. In particular, LLVQ shape--gain exhibits strong Hadamard-free PTQ behavior, indicating that sufficiently high-dimensional, low-angular-distortion codes can mitigate or even remove the need for online rotations.

From a geometric perspective, the combination of (a) high-dimensional lattice-based directional quantization and (b) norm-preserving GPTQ updates yields an effectively (hyper-)Spherical PTQ procedure: quantization primarily induces angular error, while block norms are preserved and refined via closed-form scaling. We emphasize that Spherical GPTQ is not the only reason LLVQ outperforms baselines such as E8P/Quip\# and QTIP, since even when we remove shape--gain completely and use standard (Euclidean) GPTQ, Table~\ref{tab:hadamard-ablation} shows that LLVQ still outperforms the $E_8$/Quip\# codebook. This is expected: beyond its favourable spherical-code properties, the Leech lattice also provides a lower-distortion Euclidean quantization grid than $E_8$ at comparable rates, so its advantage persists even without norm-preserving (spherical) corrections.

\subsection{Comparison to Results in Literature}

\begin{table*}[h]
\centering
\caption{
Comparison of performance after quantizing Llama-2, Llama-3, Ministral-3 and Qwen-v3 language models using different quantization methods, evaluated by Wikitext-2 (Wiki) perplexity at 4096 context length and downstream task performance (CSR and MMLU) on own consistent training pipeline. LLVQ consistently outperforms standard vector quantization approaches.}
\vspace{-0.0em}
\label{tab:other-models}
\resizebox{\linewidth}{!}{
\begin{tabular}{l r c | c c c | c c c | c c c | c c c | c c c }
& \textbf{Fine-} & 
& \multicolumn{3}{c}{Llama-2 7B}
& \multicolumn{3}{c}{Llama-3 8B}
& \multicolumn{3}{c}{Ministral-3 8B instruct}
& \multicolumn{3}{c}{Qwen-3 4B}
& \multicolumn{3}{c}{Qwen-3 8B} \\
\textbf{Method} (same pipeline) & \textbf{tuned} & \textbf{BPW}
& \textbf{Wiki}$\downarrow$ & \textbf{MMLU}$\uparrow$ & \textbf{CSR}$\uparrow$ 
& \textbf{Wiki}$\downarrow$ & \textbf{MMLU}$\uparrow$ & \textbf{CSR}$\uparrow$ 
& \textbf{Wiki}$\downarrow$ & \textbf{MMLU}$\uparrow$ & \textbf{CSR}$\uparrow$ 
& \textbf{Wiki}$\downarrow$ & \textbf{MMLU}$\uparrow$ & \textbf{CSR}$\uparrow$ 
& \textbf{Wiki}$\downarrow$ & \textbf{MMLU}$\uparrow$ & \textbf{CSR}$\uparrow$ \\
\hline
Baseline & - & 2 & 5.11 & 45.7 & 70.4 &
5.75 &
65.5 &
74.6 &
6.44 & 65.1 & 76.4 & 
12.41 & 70.2 & 71.2 & 
8.99 & 74.9 & 74.0
\\
\hline
GPTQ + Rotation (Quarot)  & No & 2 & 
41.87 & 27.0 & 41.7 &
94.37 & 25.2 & 43.3 &
41.22 & 26.7 & 44.4
& 280.7 & 26.3 & 43.6 & 41.62 & 29.9 & 47.8 \\
\cdashline{0-16}[5pt/4pt] 
Quip\#/E8P12 & No & 2 & 7.96 & 30.5 & 61.4 & 12.25 & 40.5 & 62.0 &
10.83 & 49.6 & 65.7 
& 21.15 & 48.6 & 57.2 & 
12.80 & 60.5 & 67.0 \\
QTIP (3INST) & No & 2 &
7.28 & 34.2 & 63.0 &
9.59 & \textbf{50.1} & 66.4 & 
8.96 & 55.0 & 71.6 &
17.04 & 57.4 & 63.5 &
11.17 & 66.1 & 71.1 \\
LLVQ [spherical shaping] (\textbf{ours}) & No & 2 & 7.61 & 33.4 & 62.1 &
11.49 & 41.9 & 64.8
& 10.32 & 50.2 & 66.5 & 21.80 & 50.5 & 58.7 &
12.20 & 63.7 & 68.7
 \\
LLVQ [shape-gain, 2 bit gain] (\textbf{ours}) & No & 2 & 6.83 & 34.9 & 64.6 &
9.35 & 48.7 & 66.4 &
8.56 & 56.6 & 71.3 & 
\textbf{15.54} & 59.3 & \textbf{64.1} &
10.82 &
67.2 &
69.9
\\
LLVQ [shape-gain, 0 bit gain] (\textbf{ours}) & No & 2 &
\textbf{6.48} & \textbf{35.4} & \textbf{66.5} &
\textbf{8.50} &
\textbf{52.6} &
\textbf{69.3} &
\textbf{8.11} &
\textbf{58.4} &
\textbf{72.2} &
17.05 &
\textbf{60.7} &
63.6 &
\textbf{10.19} &
\textbf{69.3} &
\textbf{70.2}
\\
\hline
Quip\#/E8P12 & Yes & 2 & 5.73 & 30.6 & 64.9 
& 7.92 & 48.1 & 66.7
& 7.54 & 54.9 & 70.6
& 10.52 & 52.9 & 65.2 &
8.31 & 63.7 & 70.1 \\
QTIP (3INST) & Yes & 2 &
5.50 & 36.9 & 66.5 &
7.28 & 53.5 & 69.1 &
7.06 & 56.7 & 73.3 &
9.61 & 59.5 & 66.9 &
7.82 & 68.0 & 72.8
\\
LLVQ [spherical shaping] (\textbf{ours}) & Yes & 2 &
5.60 &
35.8 &
65.3 &
7.64 & 47.8 & 68.7
& 7.34 & 53.8 & 70.5
& 10.13 & 54.9 & 65.1 &
8.09 & 66.4 & 71.5 \\
LLVQ [shape-gain, 2 bit gain] (\textbf{ours}) & Yes & 2 &
5.48 & 37.3 & 66.8 & 
7.29 & 53.4 & 70.0 &
7.04 & 57.6 & 72.5 &
9.51 & 60.9 & 67.6 &
7.79 & 68.8 & 72.6 \\
LLVQ [shape-gain, 0 bit gain] (\textbf{ours}) & Yes & 2 &
\textbf{5.33} & \textbf{38.0} & \textbf{68.1} &
\textbf{6.99} & \textbf{55.7} & \textbf{70.7} &
\textbf{6.83} & \textbf{59.2} & \textbf{73.0} &
\textbf{9.26} & \textbf{62.8} & \textbf{66.1} &
\textbf{7.59} & \textbf{69.6} & \textbf{72.3}
\\
\hline
\end{tabular}
}
\vspace{-2.0em}
\end{table*}

In addition to the unified PTQ evaluation presented earlier, where all methods are assessed under the exact same pipeline for strict apples‑to‑apples comparison, Table~\ref{tab:ptq-performance-literature} further broadens the analysis to include results previously reported in the literature. Although published baselines such as OmniQ, AQLM, Quip\#, and QTIP may slightly differ in training conditions, calibration set composition, and dataset sizes, these comparisons remain highly meaningful performance across independent pipelines.

Optionally, we incorporate a lightweight fine‑tuning step that learns an element‑wise multiplicative correction on the inputs of each linear layer, following \citep{tseng2024quip}; equivalently, this can be viewed as learning per‑column scaling factors for the weight matrices. Because these scalars are shared across rows, the overhead is negligible (less than $0.001$ bits per weight even in full 32‑bit precision). We train only these scale parameters for a short run of roughly 52M tokens. This minimal adaptation reliably improves perplexity, MMLU, and CSR across quantization methods, acting as a lightweight alignment step. 

Notably, LLVQ maintains a clear advantage compared to the strongest results reported across the broader literature. Crucially, LLVQ \textit{without} any fine‑tuning is competitive, sometimes surpassing, the performance of the best baselines \textit{with} fine‑tuning. This despite that we use a very strict definition of “no fine‑tuning” for our own method, meaning that all corrections arise solely from layer‑local, Hessian‑based updates derived from activation statistics, with no reliance on gradient updates, such as the inter‑layer fine‑tuning used in Quip\# and QTIP, and without any end‑to‑end tuning of the quantized model. Overall, LLVQ achieves state‑of‑the‑art performance in a strictly PTQ setting, outperforming methods that rely on additional fine‑tuning for recovery. When fine‑tuning is added (for LLVQ, only shared row-/column-wise scale terms), the performance gap widens further, yielding results close to the baseline model (2.5\%–7.6\% degradation in benchmark accuracies), pushing the frontier of practical LLM quantization into the ultra‑low‑bitrate regime of just 2 bits per weight.

\vspace{-1em}
\section{Conclusion}

We introduced \emph{Leech Lattice Vector Quantization} (LLVQ), a practical high-dimensional vector quantizer, grounded in the geometric and combinatorial structure of the Leech lattice. LLVQ provides an expressive and computationally efficient alternative to conventional scalar and low-dimensional vector quantizers. Our contributions include: (i) an extended shell-based search procedure supporting multi-shell codes, (ii) a fully invertible indexing scheme enabling codebook-free quantization and dequantization, and (iii) demonstrating Leech-lattice-based vector quantization of LLMs.

Finally, we provided a geometric interpretation of combining shape--gain with scale-corrections and GPTQ-style corrections: we show the standard scale-correction used in shape--gain acts as a retraction onto a product of spheres, yielding a \emph{Spherical GPTQ} viewpoint where the residual distortion is dominated by angular error. This perspective is particularly natural for \emph{spherical} (directional) codes, which we show can also be constructed from the Leech lattice, and empirically demonstrate reduced sensitivity to Hadamard/rotation preprocessing enabling \emph{Hadamard-free PTQ}, reducing or eliminating the need for online Hadamard transforms at inference time.

Experimentally, LLVQ achieves state-of-the-art performance in both idealized and practical settings. On Gaussian sources, LLVQ realizes the highest SQNR among competing quantizers, achieving over 92\% retention of the Shannon limit at 2~bits/dim. On all assessed large language models of the Llama-2 and Llama-3, Ministral-3, and Qwen-v3 model families, LLVQ consistently outperforms existing PTQ baselines such as AQLM, Quip\# and QTIP across perplexity and downstream task performance. This shows that the theoretical benefits of high-dimensional lattices on Gaussian data translates to practical benefits for modern LLM compression. 



Overall, LLVQ demonstrates that high-dimensional lattices offer substantial benefits for modern neural network compression. We hope this work inspires further exploration of mathematically grounded quantization schemes for scalable and efficient large model deployment.


\newpage

\bibliography{example_paper}
\bibliographystyle{icml2026}


\newpage
\newpage
\appendix
\onecolumn


\hspace{5em}

\newpage
\section{Dequantizer}
\label{sec:dequantizer}

The dequantizer recovers a $24$-dimensional integer vector from its global integer index:
{\setlength{\abovedisplayskip}{6pt}
\setlength{\belowdisplayskip}{6pt}
\setlength{\abovedisplayshortskip}{0pt}
\setlength{\belowdisplayshortskip}{2pt} 
\setlength{\jot}{3pt} 
\begin{align}
\mathrm{Dequantizer} : \{1,\dots,N(m)\} \to L^{\mathrm{int}}(m) \subset \mathbb{Z}^{24}.
\end{align}
}
Because the indexing is hierarchical, the inverse map of the dequantizer mirrors this structure and consists of a small number of inexpensive integer operations.
\vspace{-6pt}
\paragraph{1.\;Shell Identification.}
Given an index \(I\), determine the shell by locating the unique \(k\) such that
$
N(k) < I \le N(k+1).
$
This requires only a lookup in a small table of cumulative shell sizes. The shell-local index is
$
I_{\text{shell}} = I - N(k).
$
\vspace{-6pt}
\paragraph{2.\;Class Identification.}
Each shell contains a fixed, precomputed list of classes with cumulative offsets $
C_1, C_2, \dots, C_J.
$
The class index \(j\) satisfies
$
C_{j-1} < I_{\text{shell}} \le C_j,
$
and the class-local index is
$
I_{\text{class}} = I_{\text{shell}} - C_{j-1}.
$
\vspace{-6pt}
\paragraph{3.\;Unpacking local symmetries.}
Within a class, degrees of freedom factor into (i) a Golay refinement of cardinality $A$, (ii) a valid sign configuration with $2^B$ possibilities, and (iii) a permutation coset encoded by a rank in its orbit. The class-local index is unflattened by modulo and integer-division:
{\setlength{\abovedisplayskip}{4pt}
\setlength{\belowdisplayskip}{4pt}
\setlength{\abovedisplayshortskip}{0pt}
\setlength{\belowdisplayshortskip}{2pt} 
\setlength{\jot}{3pt} 
\begin{align}
r = I_{\text{class}} \bmod A, \quad \quad
I'   = \left\lfloor I_{\text{class}}/A \right\rfloor, \quad
s = I' \bmod 2^{B},               \quad
I'' = \left\lfloor I'/2^{B} \right\rfloor,
\end{align}
}
where $r$ selects the Golay refinement, $s$ selects the sign pattern, and $I''$ encodes the permutation rank.
\vspace{-6pt}
\paragraph{4. Reconstruction of the integer vector.}
Starting from the class leader, reconstruction proceeds in three conceptual stages while avoiding enumerations. First, the absolute-value pattern is rearranged by applying the permutation encoded by $I''$, yielding the correct coordinate placement for the class. Next, signs are assigned in a manner consistent with the class constraints: for \emph{odd leaders}, all $4096$ Golay codewords are admissible, and sign allocation follows the fixed $1 \bmod 4$ versus $3 \bmod 4$ structure; for \emph{even leaders}, only refinements of the appropriate Golay weight are allowed, and the sign vector must satisfy the Conway--Sloane parity and sum constraints. Finally, the Golay refinement $r$ specifies the congruence class of
$(x - m\mathbf{1})/2 \ \bmod\ 2,$
thereby fixing the remaining binary degrees of freedom and completing the integer vector in $L^{\mathrm{int}}(m)$. 
\vspace{-6pt}
\paragraph{5. Parallel Implementation (kernel).}
All components of the dequantizer, shell lookup, class lookup, and local symmetry unflattening, depend only on small static tables, integer prefix-sum scans, integer division and modulo, and local combinatorial reconstruction. There are no dependencies between vectors, no large memory accesses. Further, the procedure is therefore trivially parallelizable across blocks of 24-dimensional vectors and maps naturally to GPU execution (e.g., a CUDA kernel).

\newpage
\section{Overall of Shape-gain Quantization Procedure with Hessian Corrections}

We quantize each layer in a sequential, block-wise manner. Starting from $\widetilde{\mathbf{W}}\leftarrow\mathbf{W}^{(l)}$, we process input-channel blocks $Q_1,\dots,Q_m$ in a fixed left-to-right order. At step $t$, we apply shape--gain quantization to the current (already corrected) block $\widetilde{\mathbf{W}}_{:,Q_t}$ by quantizing its directions and then resetting the scale to match the original block norm. We form the GPTQ error $\mathbf{E}_{:,Q_t}=\widetilde{\mathbf{W}}_{:,Q_t}-\hat{\mathbf{W}}_{:,Q_t}$, which is then propagated into the remaining, unprocessed columns $R=\bigcup_{u>t}Q_u$ using the analytic Hessian correction derived in Sec.~\ref{subsec:hessian_correction}. Using the current $\widetilde{\mathbf{W}}$ (rather than the original $\mathbf{W}$) is essential: it ensures that each subsequent block is quantized only after compensating for errors introduced earlier, so the procedure greedily minimizes the local quadratic proxy at every step while keeping already-quantized blocks fixed. Interestingly, the combination of resetting the scales which is standard in shape-gain quantization and GPTQ-like updates admits a simple geometric interpretation, the scale reset enforces a constant-norm constraint on each block, so quantization primarily induces angular (directional) error on a product of spheres. The subsequent Hessian-based correction then propagates this residual angular error through the local quadratic surrogate, yielding a GPTQ-style update that effectively operates on the hypersphere rather than in unconstrained Euclidean space (see Sec.~\ref{sec:spherical-gptq}).

\begin{algorithm}[H]
\caption{Overall Shape--Gain Quantization with Hessian Corrections}
\begin{algorithmic}[1]
\Require Layer weights $\mathbf{W}^{(l)}\in\mathbb{R}^{d_{\text{out}}\times d_{\text{in}}}$, calibration inputs $\mathbf{X}\in\mathbb{R}^{N\times d_{\text{in}}}$, block size $b{=}24$, direction quantizer $\mathcal{Q}_{\text{shape}}$ (Leech), optional gain quantizer $\mathcal{Q}_{\text{gain}}$
\For{each layer $l$}
    \State $\mathbf{H}\leftarrow \frac{1}{N}\mathbf{X}^\top\mathbf{X}$ 
    \State $\mathbf{U}\leftarrow \mathrm{chol}(\mathbf{H}^{-1})^\top$ 
    \State $\widetilde{\mathbf{W}}\leftarrow \mathbf{W}^{(l)}$
    \State Partition $\{1,\dots,d_{\text{in}}\}$ into blocks $Q_1,\dots,Q_m$ of size $b$ (last may be smaller)
    \For{$t=1$ \textbf{to} $m$}
        \State $Q\leftarrow Q_t$, \quad $R\leftarrow \bigcup_{u>t} Q_u$
        \For{\textbf{each row} $i=1$ \textbf{to} $d_{\text{out}}$ \textbf{(in parallel)}}
            \State $\mathbf{v}\leftarrow \widetilde{\mathbf{W}}_{i,Q}$
            \State $\hat{\mathbf{v}}\leftarrow \|\mathbf{v}\|_2 \cdot \mathcal{Q}_{\text{dir}}\!\left(\mathbf{v}/\|\mathbf{v}\|_2\right)$
            \State \textbf{optionally:} $\hat{\mathbf{v}}\leftarrow \mathcal{Q}_{\text{gain}}(\|\mathbf{v}\|_2)\cdot \mathcal{Q}_{\text{dir}}\!\left(\mathbf{v}/\|\mathbf{v}\|_2\right)$
            \State $\hat{\mathbf{W}}_{i,Q}\leftarrow \hat{\mathbf{v}}$
        \EndFor
        \State $\mathbf{E}_{:,Q}\leftarrow \widetilde{\mathbf{W}}_{:,Q}-\hat{\mathbf{W}}_{:,Q}$
        \State $\widetilde{\mathbf{W}}_{:,Q}\leftarrow \hat{\mathbf{W}}_{:,Q}$
        \If{$R\neq \emptyset$}
            \State $\widetilde{\mathbf{W}}_{:,R}\leftarrow \widetilde{\mathbf{W}}_{:,R}-\big(\mathbf{E}_{:,Q}\mathbf{U}_{QQ}^{-1}\big)\mathbf{U}_{QR}$ 
        \EndIf
    \EndFor
    \State $\mathbf{W}^{(l)}\leftarrow \widetilde{\mathbf{W}}$
\EndFor
\end{algorithmic}
\end{algorithm}

\newpage
\section{Efficient Implementation and CUDA Kernel}
\label{sec:cuda-kernel}

\subsection{Timing results on NVIDIA with CUDA kernel}

We implement a fused CUDA kernel that performs both dequantization and dot product accumulation directly, avoiding materialization of the weight matrix and reduces memory traffic and improves throughput. This is especially important for autoregressive inference regime of large language models, where computation is typically memory-bound. We stick to the single shell $M=3$ case for simplicity.

Empirically, we observe that the fused LLVQ kernel achieves competitive performance on NVIDIA GPUs. The LLVQ kernel is faster than the FP16 matrix--vector baseline, demonstrating that the structured LLVQ representation can be evaluated efficiently without materializing the full weight matrix. While our current implementation is slightly slower than QTIP, LLVQ outperforms QTIP in terms of representation quality, as reflected by the perplexity and downstream evaluation results reported in the main paper.

\paragraph{Remark.}
We emphasize that this work does not aim to make definitive claims about optimized runtime performance, since low-level kernel engineering and hardware-specific optimizations can likely further improve the runtime of our kernel implementations. Such optimizations are largely orthogonal to the main contribution of this paper, which focuses on the quality of the representation and its impact on perplexity and downstream metrics. Nevertheless, we include these timing measurements to demonstrate the feasibility of the approach and its compatibility with efficient fused implementations.

\begin{table}[h]
   \centering
   \caption{CUDA kernel runtime for fused dequantization with matrix--vector (matvec) products}
   \resizebox{\linewidth}{!}{
   \begin{tabular}{l c l}
       \hline
       Shape & Kernel & Runtime \\
       \hline
       FP16 matvec
       & $(4096 {\times} 4096) {\cdot} (4096 {\times} 1)$ 
       & 16.3 $\mu$s \\
       FP16 matvec
       & $(4096 {\times} 4104) {\cdot} (4104 {\times} 1)$ 
       & 17.69 $\mu$s \\
       LLVQ-FP16 (fused dequant + matvec) 
       & $(4096 {\times} 4104) {\cdot} (4104 {\times} 1)$ 
       & 11.94 $\mu$s (1.36/1.48$\times$ speedup) \\
   \end{tabular}
   }
\end{table}

\newpage
\section{Shape-Gain Quantization and Constructing Spherical Codes from Lattices}
\label{sec:shape-gain}

An alternative to spherical shaping, where all lattice points within a prescribed radius are taken as the finite quantization grid, is \textit{shape–gain} quantization \citep{sabin1982product,hamkins2002gaussian,gersho2012vector}. In shape–gain, the magnitude of a vector (its “gain”) is quantized separately using a standard scalar quantizer, while its direction (its “shape”) is quantized using a spherical quantizer defined over a spherical code, i.e., a finite subset of points on the high‑dimensional unit sphere. Although this approach is slightly more involved than spherical shaping, as it requires two quantizers rather than one, it can yield superior rate–distortion performance for Gaussian sources, as we demonstrate in \autoref{app:spherical-bounding-vs-shape-gain}.

More formally, shape–gain quantization considers (polar-)decomposing each vector into its \emph{shape} (direction) and \emph{gain} (magnitude). Any non-zero vector $\mathbf{x}\in\mathbb{R}^n$ can be written as
\begin{align}
\mathbf{x} = r\,\mathbf{u},
\qquad
r=\|\mathbf{x}\| \in \R,\;\; \mathbf{u}=\mathbf{x}/\|\mathbf{x}\|\in\mathbb{S}^{n-1}.
\end{align}
Quantization then treats $r$ and $\mathbf{u}$ separately, so that the overall code becomes the Cartesian product of a scalar quantizer on $\mathbb{R}$ and a quantizer on the sphere $\mathbb{S}^{n-1}$. For Gaussian sources, this factorization is particularly appealing: the gain $r$ follows a root chi‑square distribution (since $r^2 \sim \chi^2_n$), which can be accurately quantized using an analytically tractable inverse CDF. The core design challenge therefore lies in constructing expressive, approximately uniformly distributed, and high‑rate spherical codes, together with algorithms that efficiently map a vector to its nearest code point on the sphere and represent that point using a compact index.

\paragraph{Spherical codes}

A spherical code is formally defined as a $(d, N, s)$-code: a collection of $N$ points on the sphere $S^{d-1}$ such that the inner product between any two distinct points is at most $s$. An optimal $(d, N, s)$-code is one for which no $(d, N_0, s)$-code with $N_0 > N$ exists. Among the most celebrated examples are the minimal vectors of the $E_8$ and Leech lattices, which yield an $(8, 240, \tfrac{1}{2})$-code and a $(24\; 196{,}560, \tfrac{1}{2})$-code, respectively. Their optimality was established by \citet{kabatiansky1978bounds, odlyzko1979new}, and their uniqueness by \citet{bannai1981uniqueness}. These spherical codes also solve the kissing number problem in their respective dimensions, achieving $240$ contacts in $\mathbb{R}^8$ and $196{,}560$ in $\mathbb{R}^{24}$ \citep{ericson2001codes, conway2013sphere}.

Despite their optimality, the first shells of $E_8$ and the Leech lattice contain too few points to serve as high-bitrate spherical codes: $\log_2(240)/8 \approx 1$ bit/dim for $E_8$ and $\log_2(196{,}560)/24 \approx 0.73$ bit/dim for the Leech lattice. Such bitrates are insufficient for practical neural network quantization, which typically operates in the 2--3 bits/dim or higher (up to 16) range. To obtain richer directional codebooks, we consider a normalized spherically bounded subset of the Leech lattice, i.e., the union of all lattice points within a chosen radius, which forms a dense spherical code after normalization. These multi-shell Leech-based codes serve as the foundation for our angular quantization scheme. Their geometric uniformity minimizes directional distortion, and their high density supports bitrates compatible with modern compression and quantization needs. When combined with an appropriate scalar quantizer for the gain, the resulting shape-gain quantizer performs well not only for ideal Gaussian sources but also for real-world neural network weight distributions.

\newpage
\section{Quantizer/dequantizer block diagrams}
\label{sec:qd-blocks}

We quantize a vector $\mathbf{w}\in\mathbb{R}^d$ using a mapping $q:\mathbb{R}^d\to[1,N(m)]$, where the integer index range $[1,N(m)]$ is sized such that $\log_2(N(m))/24$ equals the desired bits per dimension. The quantizer output is $i_{\hat{\mathbf{w}}}=q(\mathbf{w})$, and the reconstruction is obtained via the dequantizer $\text{Dequantizer}:[1,N(m)]\to\mathbb{R}^d$, yielding $\hat{\mathbf{w}}=\text{Dequantizer}(i_{\hat{\mathbf{w}}})$. This abstraction covers a family of encoders with geometric preprocessing (e.g., shaping or factorization) followed by index selection, and a corresponding inverse mapping at the decoder.

\paragraph{Spherical shaping.}
Figure~\ref{fig:diagram1} depicts the \emph{spherical shaping} variant based on a ball cut of the Leech lattice. The vector $\mathbf{w}$ is first projected into a spherical shaping region, typically the Euclidean ball $\mathbb{B}(0,R)$, and the index is selected by nearest-neighbor search over $\Lambda_{24}\cap\mathbb{B}(0,R)$:
\[
    i_{\hat{\mathbf{w}}}=q(\mathbf{w})
    = \arg\min_{i:\,\mathbf{c}_i\in\Lambda_{24}\cap\mathbb{B}(0,R)}\|\mathbf{w}-\mathbf{c}_i\|_2^2,
    \qquad
    \hat{\mathbf{w}}=\mathbf{c}_{i_{\hat{\mathbf{w}}}}.
\]
This realizes high shaping efficiency while enforcing a finite-energy codebook.

\begin{figure}[h!]
    \centering
    \includegraphics[width=0.80\linewidth]{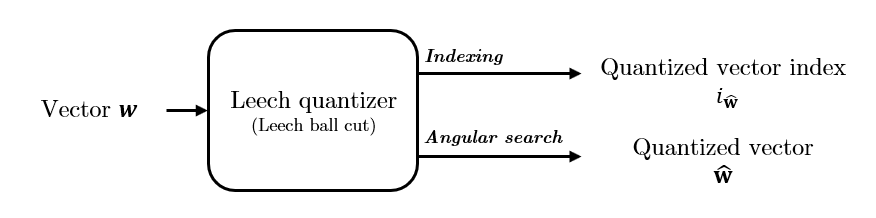}
    \caption{Spherical shaping: ball cut of the Leech lattice with nearest-neighbor selection over $\Lambda_{24}\cap\mathbb{B}(0,R)$.}
    \label{fig:diagram1}
\end{figure}

\paragraph{Shape--gain.}
In the \emph{shape--gain} framework, $\mathbf{w}=g\,\mathbf{s}$ with $g=\|\mathbf{w}\|_2$ and $\mathbf{s}=\mathbf{w}/\|\mathbf{w}\|_2\in\mathbb{S}^{d-1}$. The independent variant in Fig.~\ref{fig:diagram3} applies separate quantizers to shape and gain,
\[
    i_{\hat{\mathbf{w}}}=q(\mathbf{w})
    = \big(q_{\mathrm{shape}}(\mathbf{s}),\,q_{\mathrm{gain}}(g)\big)\in[1,N(m)],
\]
and the dequantizer reconstructs $\hat{\mathbf{s}}$ and $\hat{g}$ and returns $\hat{\mathbf{w}}=\hat{g}\,\hat{\mathbf{s}}$. While computationally simple, independence may induce norm mismatch because angular errors perturb the effective magnitude.

\begin{figure}[h!]
    \centering
    \includegraphics[width=0.80\linewidth]{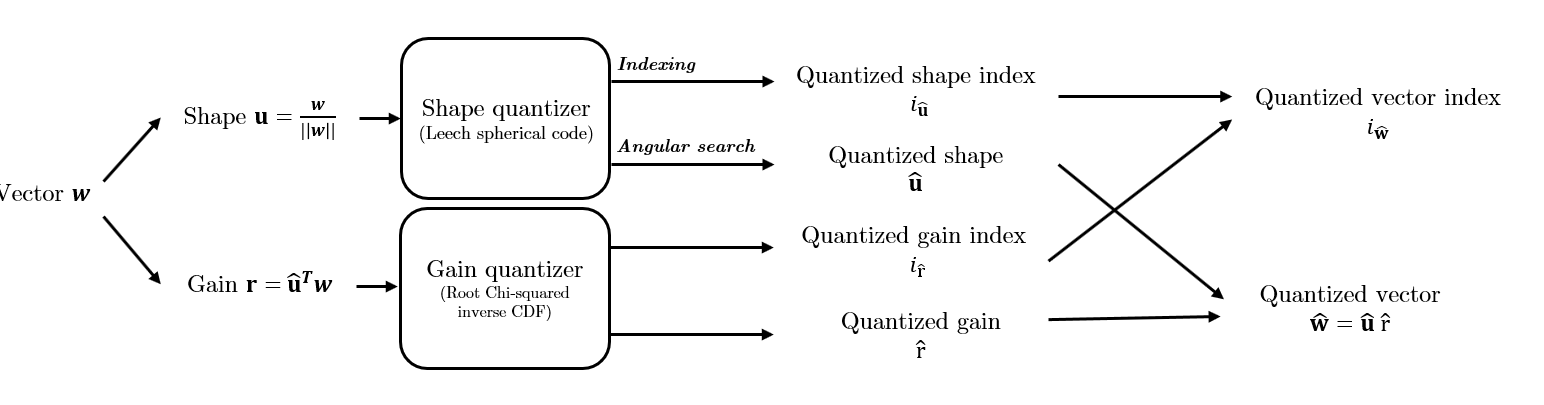}
    \caption{Shape--gain with independent quantization of shape $\mathbf{s}$ and gain $g$, producing an index in $[1,N(m)]$ and reconstruction $\hat{\mathbf{w}}=\hat{g}\,\hat{\mathbf{s}}$.}
    \label{fig:diagram3}
\end{figure}

To mitigate this, Fig.~\ref{fig:diagram2} implements \emph{shape--gain with optimal scales} (post-shape gain optimization). After selecting the shape index, the quantized direction $\hat{\mathbf{s}}$ is known; the encoder computes
\[
    \gamma^\star=\langle \mathbf{w},\hat{\mathbf{s}}\rangle,
    \qquad
    i_{\hat{\mathbf{w}}}
    = \big(q_{\mathrm{shape}}(\mathbf{s}),\,q_{\mathrm{gain}\mid\hat{\mathbf{s}}}(\gamma^\star)\big)\in[1,N(m)],
\]
and the dequantizer mirrors this conditional mapping to obtain $\hat{g}$ and reconstruct $\hat{\mathbf{w}}=\hat{g}\,\hat{\mathbf{s}}$. Unless stated otherwise, this \emph{shape--gain with optimal scales} configuration is the one used in our main LLVQ experiments due to its superior rate--distortion behavior.

\begin{figure}[h!]
    \centering
    \includegraphics[width=0.80\linewidth]{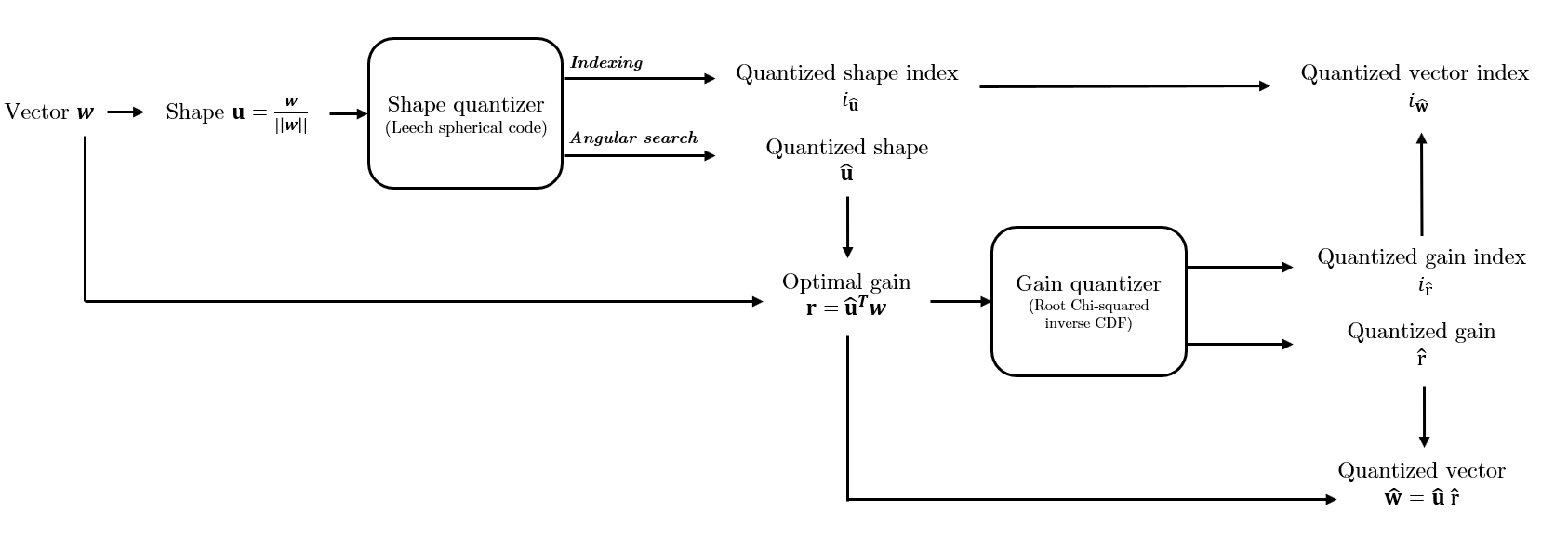}
    \caption{Shape--gain with optimal scales: post-shape gain optimization via a shape-conditioned gain quantizer.}
    \label{fig:diagram2}
\end{figure}

\paragraph{Dequantizer.}
Given the integer index $i_{\hat{\mathbf{w}}}\in[1,N(m)]$, the dequantizer in Fig.~\ref{fig:diagram4} maps back to $\mathbb{R}^d$ by retrieving the appropriate representatives (lattice point or shape/gain codewords) and recombining them:
\[
    \text{Dequantizer}:[1,N(m)]\to\mathbb{R}^d,\qquad \hat{\mathbf{w}}=\text{Dequantizer}(i_{\hat{\mathbf{w}}}).
\]
This guarantees consistency with the encoder’s shaping or factorization policy.

\begin{figure}[h!]
    \centering
    \includegraphics[width=0.80\linewidth]{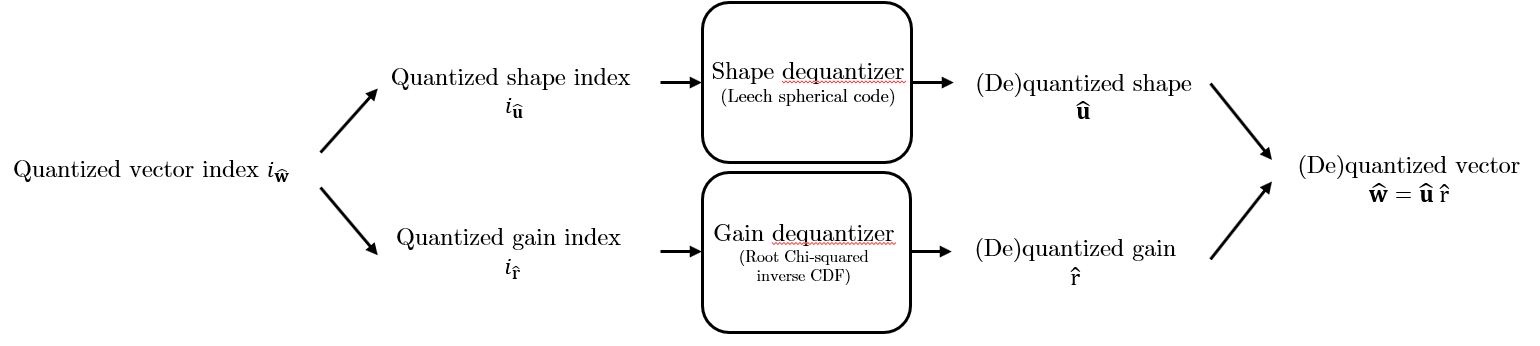}
    \caption{Dequantizer mapping the integer index $i_{\hat{\mathbf{w}}}\in[1,N(m)]$ to the reconstruction $\hat{\mathbf{w}}\in\mathbb{R}^d$.}
    \label{fig:diagram4}
\end{figure}

A detailed comparison between spherical shaping and shape--gain is provided in Appendix~\ref{app:spherical-bounding-vs-shape-gain}. There, we found shape--gain to yield improved performance over spherical shaping, when using the Leech lattice; accordingly, we adopt the shape--gain with optimal scales scheme in our main LLVQ experiments unless explicitly noted otherwise.

\newpage
\section{Algorithm specification}

\noindent
\subsection{Optimal scales under shape--gain}

A key benefit of spherical (shape--gain) vector quantization is that the \emph{shape} quantizer can be made
scale-invariant: it depends only on direction. Concretely, let
$\vu=\vw/\|\vw\|_2$ and define a spherical quantizer $q(\vw)\equiv q(\vu)$ whose codewords have fixed norm
(e.g., $\|q(\vu)\|_2=1$). Then for any $s>0$,
\[
q(s\vw)=q\!\left(\frac{s\vw}{\|s\vw\|_2}\right)=q\!\left(\frac{\vw}{\|\vw\|_2}\right)=q(\vw),
\]
i.e., $q$ is invariant to positive rescalings.\footnote{For $s<0$, directions flip. If the codebook is antipodally
symmetric (contains $\pm \vc$), then one can also enforce $q(-\vw)=-q(\vw)$. Otherwise we restrict to $s>0$.}
As a result, a scaled reconstruction of the form $\hat{\vw}(\beta)=\beta\,q(\vw/\beta)$ simplifies (for $\beta>0$) to
\[
\hat{\vw}(\beta)=\beta\,q(\vw),
\]
eliminating the need for a line search over $\beta$ and yielding closed-form optimal scales.

\paragraph{Optimal closed-form scale for weight-space error.}
With $\vq \coloneqq q(\vw)$ fixed, the scale minimizing the weight reconstruction error is
\begin{align}
\beta_* 
&= \arg\min_{\beta}\,\|\vw-\beta\,\vq\|_2^2
= \frac{\vq^\top \vw}{\vq^\top \vq},
\end{align}
i.e., the projection of $\vw$ onto its quantized shape $\vq$.
In the group-wise case, where weights are partitioned into blocks $\vw_i$, the same form applies per block:
\[
\beta_i^*=\frac{q(\vw_i)^\top \vw_i}{q(\vw_i)^\top q(\vw_i)}.
\]

\paragraph{Optimal closed-form scales for activation-space (local output) error.}
Consider a linear map with columns $\vw_i \in \mathbb{R}^{d_{\text{out}}}$ and input $x\in\mathbb{R}^G$:
\[
W x = \sum_{i=1}^G \vw_i\,x_i.
\]
Approximating $\vw_i \approx \beta_i\,q(\vw_i)$ yields
$\widehat{W}x=\sum_{i=1}^G \beta_i\,q(\vw_i)\,x_i$.
Define
\[
\mA \coloneqq \big[\,q(\vw_1)x_1,\ldots,q(\vw_G)x_G\,\big]\in\mathbb{R}^{d_{\text{out}}\times G},
\qquad \vb \coloneqq W x \in\mathbb{R}^{d_{\text{out}}}.
\]
Then the optimal group-wise scales are the least-squares solution
\begin{align}
\boldsymbol{\beta}_*
&= \arg\min_{\boldsymbol{\beta}} \|\vb-\mA\boldsymbol{\beta}\|_2^2
= (\mA^\top \mA)^{-1}\mA^\top \vb,
\end{align}
or, more robustly, with damping $\lambda>0$:
\[
\boldsymbol{\beta}_* = (\mA^\top \mA+\lambda I)^{-1}\mA^\top \vb,
\]
and for multiple calibration inputs one can stack the corresponding $(\mA,\vb)$ pairs across samples and solve the
resulting (regularized) least-squares problem.

\newpage
\subsection{Hessian-based corrections}
\label{subsec:hessian_correction}
In post-training quantization, analytic correction steps are interleaved with quantization to compensate for errors introduced by already quantized weights by adjusting the remaining ones. Instead of full fine-tuning, these corrections rely on fast analytic solutions.

\paragraph{Local Hessian Corrections (block / vector form)}
Consider a linear layer $\mathbf{y}=\mathbf{W}\mathbf{x}$ and a quantized weight matrix
$\hat{\mathbf{W}}=q(\mathbf{W})$, with error $\Delta\mathbf{W}=\hat{\mathbf{W}}-\mathbf{W}$.
A standard local proxy objective \citep{nagel2020up} is the expected output MSE
\begin{align}
\mathcal{L}_{\text{local}}
&= \mathbb{E}\big[\|\Delta\mathbf{W}\mathbf{x}\|_2^2\big]
= \mathrm{Tr}\!\big(\Delta\mathbf{W}\,\mathbf{H}_{\text{in}}\,\Delta\mathbf{W}^\top\big),
\qquad \mathbf{H}_{\text{in}}=\mathbb{E}[\mathbf{x}\mathbf{x}^\top].
\end{align}
In practice $\mathbf{H}_{\text{in}}$ is approximated by the empirical second moment
$\tilde{\mathbf{H}}_{\text{in}}=\frac{1}{N}\mathbf{X}^\top\mathbf{X}$ from a calibration set
$\mathbf{X}\in\mathbb{R}^{N\times D_{\text{in}}}$, which yields a Monte Carlo approximation that
converges to the population objective as $N\to\infty$ under i.i.d.\ sampling.

The loss decomposes over rows: $
\mathcal{L}_{\text{local}}=\sum_{r}\Delta\mathbf{w}_r^\top\mathbf{H}_{\text{in}}\Delta\mathbf{w}_r,
$ \\
so each row can be optimized independently. For a single row, suppose we quantize a \emph{set of columns}
$Q$ (possibly a block of size $|Q|>1$) and treat the remaining columns as $R$.
Partition the Hessian and the error accordingly:
\[
\mathbf{H}_{\text{in}}=
\begin{bmatrix}
\mathbf{H}_{QQ} & \mathbf{H}_{QR}\\
\mathbf{H}_{RQ} & \mathbf{H}_{RR}
\end{bmatrix},
\qquad
\Delta\mathbf{w}=[\Delta\mathbf{w}_Q;\,\Delta\mathbf{w}_R].
\]
Holding $\Delta\mathbf{w}_Q$ fixed and minimizing
$\Delta\mathbf{w}^\top\mathbf{H}_{\text{in}}\Delta\mathbf{w}$ over $\Delta\mathbf{w}_R$
gives the analytic correction
\[
\boxed{\;\Delta\mathbf{w}_R^\star=-\,\mathbf{H}_{RR}^{-1}\mathbf{H}_{RQ}\,\Delta\mathbf{w}_Q.\;}
\]
Applied row-wise, this yields the matrix update (``:’’ denotes all rows)
\[
\boxed{\;\Delta\mathbf{W}_{:,R}^\star=-\,\Delta\mathbf{W}_{:,Q}\,\mathbf{H}_{RR}^{-1}\mathbf{H}_{RQ}^\top\;}
\quad\text{(equivalently } \Delta\mathbf{W}_{:,R}^\star=-\,\Delta\mathbf{W}_{:,Q}\,\mathbf{H}_{QQ}^{-1}\mathbf{H}_{QR}\text{ under block identities).}
\]

\paragraph{Efficient implementation via inverse-Hessian Cholesky (GPTQ-style).}
Let $\mathbf{G}=\mathbf{H}_{\text{in}}^{-1}$ and compute an upper-triangular factor
$\mathbf{U}$ such that $\mathbf{G}=\mathbf{U}^\top\mathbf{U}$
(e.g., $\mathbf{U}\leftarrow\mathrm{chol}(\mathbf{G})^\top$).
When processing columns left-to-right, at any step where $Q$ precedes $R$ in the current ordering,
$\mathbf{U}$ has the block structure
$\mathbf{U}=\begin{bmatrix}\mathbf{U}_{QQ}&\mathbf{U}_{QR}\\\mathbf{0}&\mathbf{U}_{RR}\end{bmatrix}$,
and the GPTQ compensation for quantization error on $Q$ can be written without forming inverses as
\[
\boxed{\;\mathbf{W}_{:,R}\ \leftarrow\ \mathbf{W}_{:,R}\ -\ \big(\mathbf{E}_{:,Q}\,\mathbf{U}_{QQ}^{-1}\big)\,\mathbf{U}_{QR},\;}
\qquad
\mathbf{E}_{:,Q}=\mathbf{W}_{:,Q}-\hat{\mathbf{W}}_{:,Q}.
\]
This generalizes the scalar GPTQ/OPTQ update \citep{frantar2022gptq} ($|Q|=1$) to vector/block updates,
and can be interpreted as a Schur-complement / conditioning step in the local quadratic.

\paragraph{Question: Are local proxy objectives problematic?}

The local proxy objective above minimizes the expected \emph{layer-output} distortion
$\mathbb{E}\|(\hat{\mathbf{W}} - \mathbf{W})\mathbf{x}\|_2^2$ using only the input second moment $\mathbf{H}_{\text{in}}=\mathbb{E}[\mathbf{x}\mathbf{x}^\top]$
\citep{nagel2020up}. From a second-order perspective, this corresponds to a very coarse approximation of the
\emph{global} curvature of the end-to-end loss: (i) it effectively treats layers as decoupled (a block-diagonal
approximation across layers, ignoring cross-layer coupling and error propagation), and (ii) within each layer it
retains only the \emph{input-side} curvature while discarding loss-dependent \emph{output-side} weighting and the
effects of downstream nonlinearities. Equivalently, it can be viewed as a Kronecker-/Gauss--Newton-like surrogate
that keeps $\mathbf{H}_{\text{in}}$ but replaces the output factor by an identity (or constant) term, thereby ignoring
how sensitive different output directions are under the task loss.

More faithful curvature surrogates that incorporate loss information and/or cross-layer interactions (e.g.,
approximations to the global Hessian or related second-order structure) can yield stronger compression performance,
but typically require backward passes and higher calibration cost \citep{van2023llm,tseng2025model}. Since our
focus in this work is on the \emph{representation/quantizer} rather than on increasingly powerful correction
procedures, we evaluate all methods under the same local, GPTQ-like correction scheme (generalized to vector
updates) to isolate gains attributable to the representation \citep{frantar2022gptq}. Global-Hessian-based
corrections are orthogonal to our contribution, and combining them with our representation is a natural direction
for future work \citep{van2023llm,tseng2025model}.

%

\newpage
\section{Question: Is it better to construct spherical codes from a single Leech shell or a union of shells?}
\label{app:union-of-shells}

High–dimensional spherical codes derived from the Leech lattice can be constructed in multiple ways, depending on how lattice points are selected and normalized. In practice, two natural options arise: selecting points from a \emph{single} Leech shell, or taking the \emph{union of multiple shells} up to a radius threshold. Since these constructions differ in both cardinality growth and geometric diversity, it is not immediately clear which yields better angular uniformity per bit. Here, we empirically compare the two. Given a spherical code constructed from the Leech lattice in \(\mathbb{R}^{24}\), is it better (in terms of angular uniformity per bit) to take a \emph{single} shell \(m\) or to take the \emph{union of all shells up to \(m\)}?

For a finite set of unit vectors \(\mathcal{C} \subset \mathbb{S}^{23}\), we define:
\[
\text{bits} = \log_2|\mathcal{C}|,\qquad 
q(x) = \arg\min_{y \in \mathcal{C}\setminus\{x\}} \arccos (x^{\top} y ) / \pi,
\]
where \(q(x)\) is the nearest neighbor of \(x\) in \(\mathcal{C}\) under angular distance. We measure angular distance \(D_{\mathbb{S}^{23}} : \mathbb{R}^{24} \times \mathbb{R}^{24} \to [0, 1]\), shorthand \(D\), between a point and its closest radial neighbour:
\[
D(x, q(x)) = \arccos (x^{\top} q(x)) / \pi,
\]
the angular distance between \(x\) and its nearest neighbor. We report the distribution of \(D(x, q(x))\) after sampling \(x\) from a radially uniform source (such as \(x \sim \mathcal{N}(0,\mI)\) normalized to unit norm).

\begin{figure}[h!]
  \centering
  \resizebox{0.4\linewidth}{!}{
  \includegraphics[]{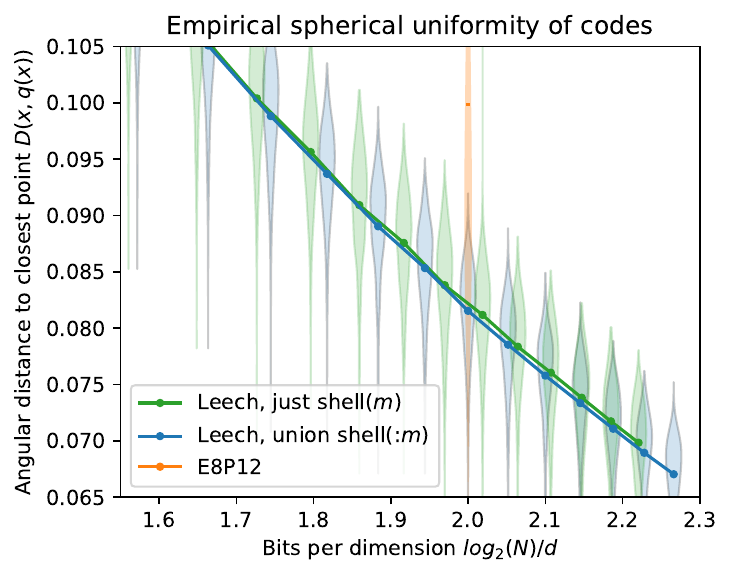}
  }
  \caption{Empirical angular separation vs.\ rate. For each code, we show the distribution over nearest–neighbor angle against bits per dimension \(\log_2(N)/d\). The two Leech-based constructions are traced by varying \(m\) (single shell \(m\) vs.\ union \(1{:}m\)). We compare to E8P12 and find that it lies above the Leech curves at the same bits (worse angular separation), while between the Leech variants the union has slightly superior angular distance.}
  \label{fig:leech-shells-vs-union}
\end{figure}

We evaluate three code families as a function of bits: (1) Leech (single shell \(m\)), i.e., points from a single Leech shell projected to the unit sphere; (2) Leech (union up to \(m\)), i.e., the cumulative union of shells \(1{:}m\), all normalized to unit norm; and (3) the E8P12 product code for reference, normalized to the unit sphere and stacked three times to allow comparison in 24 dimensions. Figure~\ref{fig:leech-shells-vs-union} shows violin plots of the resulting distributions over \(D_{24}(x, q(x))\) versus bits for these three constructions. Empirically, we find that both Leech constructions, single shell \(m\) and union up to shell \(m\), perform similarly, with the union exhibiting slightly better code quality per bit as measured by angular uniformity.

\textbf{Key finding 1: Union of shells provide lowest angular distortion.} Our empirical results indicate that spherical codes formed by normalizing a union of Leech lattice shells yields a slightly better Gaussian rate--distortion curves compared to forming spherical codes from individual Leech lattice shells. We therefore adopt this approach in our method and recommend doing the same.

\textbf{Key finding 2: Single shell provides a simpler algorithm.} Although a union of shells results in slightly lower angular distortion, the difference between the two is small. From a hardware perspective, however, using a single shell offers significant advantages. In particular, a constant norm implies a fixed scaling between dot products, eliminating the need to rescale intermediate dot product results before aggregation (as in group-wise or block quantization), along with its associated complications.

\newpage
\section{Question: which has lowest distortion: `spherical shaping' or `shape–gain'?}
\label{app:spherical-bounding-vs-shape-gain}

We consider two natural ways to construct a finite representation from a finite set of spherically bounded Leech lattice points (or from a single lattice shell), namely \emph{spherical shaping}, which uses the unnormalized points directly, and \emph{shape–gain}, which separates direction and magnitude via a projected spherical code for shape and a scalar gain code. Define \(C_{\text{spherical-shaping}} = \Lambda_{24}(m)\), the set of all lattice points in \(\Lambda_{24}\) whose squared norm is at most \(m\). We keep every lattice point inside a ball, and the resulting codebook directly inherits both its directions and its radii from the geometry of \(\Lambda_{24}\), up to a channel-wise or global scale. Intuitively, this is a very simple construction: one hyperparameter \(m\) determines both how far the code extends in radius and how many points it contains. The tradeoff between shape and gain is fixed implicitly by the lattice itself.

For shape–gain, we instead separate a vector into its magnitude and direction, \(x = r\,u\) with \(u = x/\|x\|\) and \(r=\|x\|\), and quantize these two pieces independently. Formally, define the shape code \(C_{\text{shape}} = \{\, x/\|x\| : x \in \Lambda_{24}(m)\,\}\), i.e., lattice points normalized to lie on the unit sphere, which gives a dense spherical code on \(S^{23}\). The gain is represented by a scalar gain code \(C_{\text{gain}}\) (e.g., matched to a \(\chi_{24}\) distribution for Gaussian sources). The full shape–gain code is then \(C_{\text{shape-gain}} = \{\, \hat r\, \hat u : \hat u \in C_{\text{shape}},\ \hat r \in C_{\text{gain}}\,\} = C_{\text{gain}} \times C_{\text{shape}}\). Intuitively, spherical shaping uses the raw lattice: you pick a lattice vector and both its direction and its length are “baked in.’’ In contrast, shape–gain forces every codeword direction to live on the sphere (from the projected lattice) and assigns its radius explicitly through the gain quantizer.

Because shape–gain has two components, it requires a choice of how to allocate bits between the direction (shape) part and the radius (gain) part. High-resolution theory provides a useful heuristic: in \(n\) dimensions, about \(1/n\) of the total bits should be devoted to gain, meaning that for \(n=24\) one should allocate roughly \(1/24\) of the rate to \(C_{\text{gain}}\) and the remainder to \(C_{\text{shape}}\). This gives a principled starting point and greatly reduces the effective search over hyperparameters. In contrast, the spherically bounded code \(C_{\text{spherically-bounded}} = \Lambda_{24}(m)\) has only one hyperparameter \(m\), which makes it easier to sweep but less expressive: it cannot optimize radius resolution independently of directional resolution.

We therefore compare these two constructions empirically under matched rate budgets: (i) the spherically bounded approach using \(C_{\text{spherically-bounded}} = \Lambda_{24}(m)\), and (ii) the shape–gain approach using \(C_{\text{shape-gain}} = C_{\text{gain}} \times C_{\text{shape}}\), where \(C_{\text{shape}} = \{x/\|x\| : x \in \Lambda_{24}(m)\}\) and \(C_{\text{gain}}\) is a \(\chi_{24}\)-matched scalar quantizer with approximately \(1/24\) of the available bits. This allows us to isolate the effect of separating radius and direction from the effect of simply using a spherically truncated lattice. 

\begin{table}[h]
    \centering
\resizebox{\linewidth}{!}{
    \begin{tabular}{l r c c | c c c}
       \textbf{Method} & \textbf{Code} &  & \textbf{Bits/dim} & \textbf{$\widehat{\text{MSE}}$} & \textbf{$\widehat{\text{SQNR}}$} (bits) & \textbf{Ret}(\%) \\
       \hline
       Leech (spherical bounding) & $\Lambda_{24}(13) $ & & 2.0 & 0.084 & 1.787 & 89.37 \\
       \hline
       Leech (shape-gain) & $\text{norm}(\Lambda_{24}(13))$ + 0 $\chi$-gain bits & 2.00000 + 0.00000 & 2.0 & 0.085 & 1.782 & 89.12 \\
       Leech (shape-gain) & $\text{norm}(\Lambda_{24}(12))$ + 1 $\chi$-gain bits & 1.95833 + 0.04167 & 2.0 & 0.078 & 1.843 & \textbf{92.14} \\
       Leech (shape-gain) & $\text{norm}(\Lambda_{24}(11))$ + 2 $\chi$-gain bits & 1.91667 + 0.08333 & 2.0 & 0.080 & 1.825 & 91.24 \\
       Leech (shape-gain) & $\text{norm}(\Lambda_{24}(10))$ + 4 $\chi$-gain bits & 1.83333 + 0.16667 & 2.0 & 0.085 & 1.780 & 89.01 \\
       \hline
    \end{tabular}
    }
    \caption{Comparison between spherical shaping and shape--gain quantization at a fixed budget of 2 bits/dim. For shape--gain we vary the allocation between directional (shape) bits and radial (gain) bits while keeping the total rate fixed.}
    \label{tab:spherical-bounding-vs-shape-gain}
    \vspace{-2em}
\end{table}


Taken together, the results in Table~\ref{tab:spherical-bounding-vs-shape-gain} reveal several consistent trends regarding the relative behaviour of spherical shaping and the shape--gain construction at a fixed rate of 2 bits/dim. In particular, although spherical shaping provides a clean geometric baseline, the shape--gain decomposition offers clear advantages in both distortion and retention.
\begin{itemize}
    \item \textbf{Key finding 1.} Shape--gain can improve performance over spherical shaping. 
    \item \textbf{Key finding 2.} The $1/n$ gain‑bit heuristic is reasonable but not always optimal at finite rate. High‑resolution theory in $n$ dimensions suggests allocating approximately $1/n$ of the bits to the gain; for $n=24$ this corresponds to 2 gain bits (0.083 bits/dim), whereas our empirical optimum is 1 gain bit (0.041 bits/dim).
    \item \textbf{Key finding 3.} For the Leech‑lattice shape–gain code at 2 bits/dim, we recommend using 1 or 2 gain bits per vector of length 24 (0.041-0.08333 bits/dim), though practitioners should consider running similar sweeps when targeting other bitrates to obtain the best quantization performance.
\end{itemize}

\newpage

\newpage
\section{Spherical GPTQ and Rotation/Hadamard-free PTQ}
\label{sec:spherical-gptq}

This appendix provides full derivations and pseudocode for \emph{Spherical GPTQ}, which is introduced in the main paper in \autoref{sec:spherical-gptq-main}. Applying vector or shape--gain quantization inside GPTQ immediately raises a practical question: \emph{what do we do with magnitudes?}
Given a weight block (or row/column vector) $w$, a shape--gain view decomposes it into a direction and a gain, $w = s u$ with $u \in \mathbb{S}^{d-1}$ and $s=\|w\|_2$.
After quantizing the direction (the ``shape''), there are multiple reasonable choices for the gain before also quantizing it: (i) keep the original gain ($s=\|w\|_2$),
(ii) choose $s$ to minimize the post-quantization MSE (optionally Hessian-weighted), and
(a) quantize $s$ immediately per GPTQ block, or
(b) keep $s$ in high precision during GPTQ and only quantize/solve for it after all blocks are processed.
While these choices are often treated as engineering details, in a Hessian-aware procedure like GPTQ they \emph{change the effective update rule}: the residual corrections depend on the magnitude of previously-quantized blocks, hence different gain policies induce different error-feedback dynamics.

\paragraph{Vector GPTQ (Euclidean)}
Generalizing GPTQ to vector quantization can be viewed as approximately solving, block-by-block or coordinate-by-coordinate,
\begin{equation}
\min_{\hat w \in \mathcal{C}}
\frac{1}{2}(w-\hat w)^\top H (w-\hat w),
\label{eq:gptq_objective}
\end{equation}
by greedily quantizing a subset of coordinates/blocks and propagating the resulting error to the remaining ones using a Cholesky factor of $H^{-1}$.
Algorithm~1 (Vector GPTQ) instantiates this Euclidean mechanism: after quantizing the current coordinate (or vector chunk), the residual is updated by a Hessian-preconditioned correction on the yet-unquantized coordinates.

When the quantizer $\mathcal{Q}$ is primarily \emph{directional} (e.g., shape--gain codes or lattice-based VQ used as a spherical code), the Euclidean GPTQ inner loop can exhibit an important interaction with the Hessian-weighted error-feedback updates. Although quantization always induces both angular and radial error, we empirically find that the \emph{radial component} can become disproportionately large across greedy block steps.

\paragraph{Spherical GPTQ (Riemmanian)}
Motivated by the standard ``gain reset'' step in classical shape--gain quantization, where one quantizes direction and then matches the original norm, we introduce a simple drop-in modification to the GPTQ loop: after quantizing a block, we \emph{retract} the candidate back to the original constant-norm manifold (keeping high-precision), preserving direction while eliminating norm error.
For a vector $w$ and its quantized candidate $\tilde w=\mathcal{Q}(w)$, the spherical retraction is
\begin{equation}
\hat w_{\mathrm{sph}}
\;=\;
\Pi_{\|w\|}(\tilde w)
\;:=\;
\frac{\|w\|_2}{\|\tilde w\|_2}\,\tilde w.
\label{eq:spherical_projection}
\end{equation}
In our block setting, quantization is performed row-wise on each row-block $W_{i,B}\in\mathbb{R}^{|B|}$, and we apply the same retraction per row:
\[
\hat W_{i,B}
\;=\;
\frac{\|W_{i,B}\|_2}{\|\tilde W_{i,B}\|_2}\,\tilde W_{i,B}.
\]
Geometrically, \eqref{eq:spherical_projection} is a \emph{retraction} onto a product of spheres (one sphere per row-block), turning the otherwise Euclidean GPTQ inner loop into a \emph{Riemannian} step: quantization selects a direction (a point in the tangent space proxy), and the retraction maps it back onto the fixed-norm manifold.
This motivates the name \textit{Spherical GPTQ}.
Importantly, the retraction is codebook-agnostic: any quantizer that produces a candidate $\tilde w$ can be followed by $\Pi_{\|w\|}(\cdot)$ as a drop-in modification.

Operationally, this corresponds to keeping gains in high precision during the GPTQ loop: we quantize primarily \emph{directions}, set the block norm to match the original, and then compute the GPTQ error using the retracted weights.
Concretely, for each block $B$ we quantize to obtain $\tilde R_{:,B}$, retract row-wise to $R_{:,B}$ as above, form the residual $E_{:,B}=W_{:,B}-R_{:,B}$, and propagate this residual to the remaining blocks using the standard GPTQ correction.

After all blocks are quantized (directions fixed), we optionally refine magnitudes by solving for optimal gains (\emph{group scales}) in closed form.
The post-processing step in Algorithm~2 forms, for each row $i$, a small linear system over blocks with
\[
M_{i,pq} = R_{i,B_p} H_{B_pB_q} R_{i,B_q}^\top,
\qquad
r_{i,p} = R_{i,B_p} H W_{i,:}^\top,
\]
and solves $M_i s_i = r_i$, followed by $R_{i,B}\leftarrow s_{i,B}R_{i,B}$.
Conceptually, this computes the Hessian-weighted best gains given fixed directions: we set directions via (spherical) GPTQ, then perform a single global gain update at the end.

\subsection{Pseudocode for Vector GPTQ (Euclidean) and Spherical GPTQ (Riemmanian)}

\begin{algorithm}[H]
\caption{Vector GPTQ (Euclidean)}
\begin{algorithmic}[1]
\Require $\mathbf{W}\in\mathbb{R}^{k\times d}$, SPD $\mathbf{H}\in\mathbb{R}^{d\times d}$, blocks $Q_1,\dots,Q_m$ (processing order), quantizer $\mathcal{Q}$
\State $\mathbf{U}\leftarrow \mathrm{chol}(\mathbf{H}^{-1})^\top$ \Comment{$\mathbf{H}^{-1}=\mathbf{U}^\top\mathbf{U}$, $\mathbf{U}$ upper-triangular}
\State $\widetilde{\mathbf{W}}\leftarrow \mathbf{W}$
\For{$t=1$ \textbf{to} $m$}
    \State $Q\leftarrow Q_t$, \quad $R\leftarrow \bigcup_{u>t} Q_u$
    \State $\hat{\mathbf{W}}_{:,Q}\leftarrow \mathcal{Q}(\widetilde{\mathbf{W}}_{:,Q})$
    \State $\mathbf{E}_{:,Q}\leftarrow \widetilde{\mathbf{W}}_{:,Q}-\hat{\mathbf{W}}_{:,Q}$
    \State $\widetilde{\mathbf{W}}_{:,Q}\leftarrow \hat{\mathbf{W}}_{:,Q}$
    \If{$R\neq \emptyset$}
        \State $\widetilde{\mathbf{W}}_{:,R}\leftarrow \widetilde{\mathbf{W}}_{:,R}-\big(\mathbf{E}_{:,Q}\mathbf{U}_{QQ}^{-1}\big)\mathbf{U}_{QR}$
        \Comment{right-side triangular solve in $\mathbf{U}_{QQ}$}
    \EndIf
\EndFor
\State \Return $\hat{\mathbf{W}}\leftarrow \widetilde{\mathbf{W}}$
\end{algorithmic}
\end{algorithm}

\begin{algorithm}[H]
\caption{Spherically Projected GPTQ with Group-scales (Riemannian)}
\begin{algorithmic}[1]
\Require $\mathbf{W}\in\mathbb{R}^{k\times d}$, SPD $\mathbf{H}\in\mathbb{R}^{d\times d}$, blocks $Q_1,\dots,Q_m$, quantizer $\mathcal{Q}$, damping $\lambda\ge 0$
\State $\mathbf{U}\leftarrow \mathrm{chol}(\mathbf{H}^{-1})^\top$
\State $\widetilde{\mathbf{W}}\leftarrow \mathbf{W}$
\For{$t=1$ \textbf{to} $m$}
    \State $Q\leftarrow Q_t$, \quad $R\leftarrow \bigcup_{u>t} Q_u$
    \State $\widetilde{\mathbf{W}}_{:,Q}\leftarrow \frac{\|\mathbf{W}_{:,Q}\|_F}{\|\mathcal{Q}(\widetilde{\mathbf{W}}_{:,Q})\|_F}\,\mathcal{Q}(\widetilde{\mathbf{W}}_{:,Q})$
    \State $\mathbf{E}_{:,Q}\leftarrow \mathbf{W}_{:,Q}-\widetilde{\mathbf{W}}_{:,Q}$
    \If{$R\neq \emptyset$}
        \State $\widetilde{\mathbf{W}}_{:,R}\leftarrow \widetilde{\mathbf{W}}_{:,R}-\big(\mathbf{E}_{:,Q}\mathbf{U}_{QQ}^{-1}\big)\mathbf{U}_{QR}$
    \EndIf
\EndFor

\Statex \Comment{Final per-row block scale refinement in the Hessian metric}
\For{$i=1$ \textbf{to} $k$}
    \State $\mathbf{M}_i[p,q]\leftarrow \widetilde{\mathbf{W}}_{i,Q_p}\,\mathbf{H}_{Q_pQ_q}\,\widetilde{\mathbf{W}}_{i,Q_q}^\top \quad \forall\,p,q\in\{1,\dots,m\}$
    \State $\mathbf{r}_i[p]\leftarrow \widetilde{\mathbf{W}}_{i,Q_p}\,\mathbf{H}_{Q_p,:}\,\mathbf{W}_{i,:}^\top \quad \forall\,p\in\{1,\dots,m\}$
    \State $\mathbf{s}_i\leftarrow (\mathbf{M}_i+\lambda \mathbf{I})^{-1}\mathbf{r}_i$
    \For{$p=1$ \textbf{to} $m$}
        \State $\widetilde{\mathbf{W}}_{i,Q_p}\leftarrow \mathbf{s}_i[p]\;\widetilde{\mathbf{W}}_{i,Q_p}$
    \EndFor
\EndFor
\State \Return $\hat{\mathbf{W}}\leftarrow \widetilde{\mathbf{W}}$
\end{algorithmic}
\end{algorithm}

\newpage
\subsection{Ablation: Ablating need for Hadamard rotations under Spherical GPTQ}

\begin{table}[h]
\centering
\caption{Extended ablation study with and without Hadamard rotations, evaluated on wikitext-2 perplexity (PPL) and downstream tasks (CSR and MMLU). LLVQ consistently outperforms standard VQ approaches in performance against bits per weight (BPW).}
\label{tab:hadamard-ablation-extended}
\resizebox{1.0\linewidth}{!}{
\begin{tabular}{l | c c l | c c | c c c }
& & & & 
& \textbf{Effective} & \multicolumn{3}{c}{Llama-2 7B} \\
\textbf{Method} \small{(all results in this table are without finetuning)} & \textbf{Dim} &
\footnotesize{\textbf{BPW}} & 
\textbf{Hadamard} & & \textbf{Correction}
& \textbf{Wiki} $\downarrow$ & \textbf{MMLU} $\uparrow$ & \textbf{CSR} $\uparrow$ \\
\hline
Baseline & 1 
& 16 & - & & & 5.12 & 45.7 & 70.4 \\
\hline
Integer (GPTQ) & 1 
& 2
& No Rotation & 
& GPTQ &
3411.6 & 26.6 & 39.7 \\
Integer (Quarot) & 1 & 
2 & 
Input &
& GPTQ &
41.87 & 27.0 & 41.7 \\
Integer & 1 
&
2 &
\small{Input + Output} &
& GPTQ &
37.83 & 26.1 & 48.4
\\
\cdashline{0-8}[5pt/4pt] 
E8P & 8 &
2 &
No Rotation &
& GPTQ &
105.98 & 24.8 & 44.9 \\
E8P & 8 & 
2 &
Input & 
& GPTQ &
9.24 & 31.0 & 59.8 \\
E8P (Quip\#) & 8 &
2 &
\small{Input + Output} &
& GPTQ &
7.96 & 30.5 & 61.4
\\
\cdashline{0-8}[5pt/4pt] 
\cdashline{0-8}[5pt/4pt] 
\cdashline{0-8}[5pt/4pt] 
LLVQ [spherical shaping] 
& 24 & 2 &
No Rotation &
& GPTQ &
191.90 & 24.0 & 53.5 \\
LLVQ [spherical shaping] 
& 24 & 2 & 
Input &
& GPTQ &
\textbf{6.80} & \textbf{35.1} & \textbf{65.4}
 \\
LLVQ [spherical shaping] 
& 24 & 2 &
\small{Input + Output} &
& GPTQ &
\textbf{7.61} & \textbf{33.4} & \textbf{62.1} \\
\cdashline{0-8}[5pt/4pt] 
LLVQ [spherical shaping] (forced angular correction) 
& 24 & 2 & 
No Rotation &
& \small{Spherical GPTQ} &
\textbf{6.90} & \textbf{37.4} & \textbf{63.8} \\
LLVQ [spherical shaping] (forced angular correction) 
& 24 & 2 &
Input & 
& \small{Spherical GPTQ} &
\textbf{6.70} & \textbf{35.1} & \textbf{65.4}
\\
LLVQ [spherical shaping] (forced angular correction) 
& 24 & 2 &
\small{Input + Output} &
& \small{Spherical GPTQ} &
\textbf{6.75} & \textbf{36.9} & \textbf{63.8} 
 \\
\cdashline{0-8}[5pt/4pt] 
LLVQ [shape-gain w/ 2 bit gain] (forced Euclidean correction) 
& 24 & 2 & 
No Rotation & 
& GPTQ & 
13.17 & 26.5 & 58.5
\\
LLVQ [shape-gain w/ 2 bit gain] (forced Euclidean correction) 
& 24 & 2 &
Input & 
& GPTQ &
\textbf{7.28} & 
\textbf{34.1} & 
\textbf{62.8}
\\
LLVQ [shape-gain w/ 2 bit gain] (forced Euclidean correction) 
& 24 & 2 &
\small{Input + Output} &
& GPTQ &
\textbf{7.31} &
\textbf{35.3} &
\textbf{62.8}
 \\
\cdashline{0-8}[5pt/4pt] 
LLVQ [shape-gain w/ 2 bit gain] 
& 24 & 2 & 
No Rotation &
& \small{Spherical GPTQ} &
\textbf{7.27} & \textbf{29.8} & \textbf{61.5} \\
LLVQ [shape-gain w/ 2 bit gain] 
& 24 & 2 &
Input & 
& \small{Spherical GPTQ} &
\textbf{6.90} & \textbf{36.0} & \textbf{63.6}
\\
LLVQ [shape-gain w/ 2 bit gain] 
& 24 & 2 &
\small{Input + Output} & 
& \small{Spherical GPTQ} &
\textbf{6.83} & \textbf{34.9} & \textbf{64.6}
 \\
\hline
\end{tabular}
}
\end{table}

Empirically, we find that:
\begin{enumerate}
    \item \textbf{Key finding 1. Spherical GPTQ can improve results over Euclidean GPTQ.}  
    Spherical GPTQ serves as a simple drop-in modification to standard GPTQ that improves quantization quality by explicitly removing radial error through norm preservation. By projecting quantized vectors back onto the constant-norm manifold, it ensures that residual corrections focus on the most relevant degrees of freedom. Empirically, this can lead to improved performance across quantizers without requiring changes to the underlying codebook, demonstrating that spherical projection itself is a robust and broadly applicable enhancement to GPTQ.

    \item \textbf{Key finding 2. Spherical GPTQ is most effective with low angular distortion codes.}  
    Empirically, we find that quantizers with low angular distortion (such as LLVQ) achieve the best overall performance. Notably, improvements correlate more strongly with reduced angular distortion than with low MSE under a Gaussian source model. We find that allocating bits to gain becomes largely unnecessary: capacity is better spent entirely on encoding directions. We observe an the empirical shift from using explicit gain quantization (e.g., 2-bit gains under Euclidean GPTQ) to effectively zero-bit gains under Spherical GPTQ, since norm preservation combined with low angular distortion makes additional gain resolution redundant.

    \item \textbf{Key finding 2. LLVQ outperforms under both Euclidean GPTQ and Spherical GPTQ.} LLVQ achieves the best accuracy--compression trade-offs compared to baselines regardless of whether the correction step is Euclidean (standard GPTQ) or spherical, indicating that its gains stem from the representation rather than a particular correction heuristic. The strong results under \emph{forced} mismatched corrections further suggest that LLVQ is robust to the specific correction dynamics used.

    \item \textbf{Key finding 3. Diminishing returns from Hadamard rotations.}
    With spherical corrections, the incremental benefit of Hadamard preprocessing is substantially reduced: once radial errors are eliminated and the residual updates focus on directions, the remaining distortion is dominated by the intrinsic angular resolution of the codebook rather than by coordinate anisotropy that rotations would mitigate.

    \item \textbf{Key finding 4. Bit allocation shifts toward directions (``0-bit gains'').}
    When the underlying quantization grid has low angular distortion (e.g., strong vector/lattice codes), spherical GPTQ makes allocating bits to gains less attractive.
    In this regime, we observe that the optimal shape--gain budget shifts from explicit gain quantization (e.g., ``2-bit gains'') toward effectively \emph{no gain bits}: all capacity is used to encode directions, while magnitudes are handled by the high-precision spherical constraint during GPTQ and a single closed-form Hessian update afterward.
\end{enumerate}

\newpage

\newpage
\section{Additional results on smaller Llama-3.2 1B and Llama-3.2 3B models}

\begin{table*}[h]
\centering
\caption{
Comparison of performance after quantizing smaller Llama-3.2 1B and Llama 3.2 3B language models using different quantization methods, evaluated by Wikitext-2 (Wiki) perplexity at 4096 context length and downstream task performance (CSR and MMLU) on own consistent training pipeline. LLVQ consistently outperforms standard vector quantization approaches.}
\vspace{-0.0em}
\label{tab:other-models}
\resizebox{1.0\linewidth}{!}{
\begin{tabular}{l r c c | c c c | c c c }
& \textbf{Fine-} &  &
& \multicolumn{3}{c}{Llama-3.2 1B}
& \multicolumn{3}{c}{Llama-3.2 3B}
\\
\textbf{Method} (same pipeline) & \textbf{tuned} & \textbf{BPW} & \textbf{Hadamard} 
& \textbf{Wiki}$\downarrow$ & \textbf{MMLU}$\uparrow$ & \textbf{CSR}$\uparrow$ 
& \textbf{Wiki}$\downarrow$ & \textbf{MMLU}$\uparrow$ & \textbf{CSR}$\uparrow$ 
\\
\hline
Baseline & - & 2 & \small{Input + Output} &
\textbf{12.14} & \textbf{46.2} & \textbf{61.8} &
\textbf{10.48} & \textbf{59.8} & \textbf{67.6}
\\
\hline
GPTQ + Rotation (Quarot)  & No & 2 & \small{Input + Output} &
1158.7 & 24.2 & 36.9 &

\\
\cdashline{0-9}[5pt/4pt] 
E8P/Quip\# (Euclidean GPTQ) & No & 2 & \small{Input + Output} &
38.07 & 26.4 & 49.4 &
30.77 & 36.4 & 54.9 
\\
E8P (+Spherical GPTQ) & No & 2 & \small{Input + Output} &
24.21 & 31.29 & 52.1 &
20.34 & 42.3 & 58.3
\\
%
LLVQ (spherical shaping, Euclidean GPTQ) (\textbf{ours}) & No & 2 & \small{Input + Output} &
25.36 &
33.1 &
53.1 &
18.40 &
46.0 &
60.5
 \\
LLVQ (spherical shaping, Spherical GPTQ) (\textbf{ours}) & No & 2 & \small{Input + Output} &
23.76 &
30.2 &
53.5 &
19.70 &
47.1 &
59.2
 \\
LLVQ (shape-gain w/ 0 bit gain, Euclidean GPTQ) (\textbf{ours}) & No & 2 & \small{Input + Output} &
47.78 &
25.1 &
47.5 &
32.20 &
37.9 &
52.0
\\
LLVQ (shape-gain w/ 0 bit gain, Spherical GPTQ) (\textbf{ours}) & No & 2 & \small{Input + Output} &
\textbf{21.36} & 
\textbf{32.5} &
\textbf{54.1} &

\textbf{15.93} &
\textbf{48.6} &
\textbf{60.0}
\\
LLVQ (shape-gain w/ 2 bit gain, Euclidean GPTQ) (\textbf{ours}) & No & 2 & \small{Input + Output} &
26.85 &
27.7 &
51.3 &
22.96 &
44.8 &
58.2 
\\
LLVQ (shape-gain w/ 2 bit gain, Spherical GPTQ) (\textbf{ours}) & No & 2 & \small{Input + Output} &
23.08 &
31.1 &
52.4 &
18.81 & 45.6 & 59.1
\\
\hline
E8P/Quip\# & Yes & 2 & \small{Input + Output} &
16.67 & 26.5 & 52.0 &
11.13 & 42.1 & 61.0
\\
%
LLVQ (spherical shaping, Euclidean GPTQ) (\textbf{ours}) & Yes & 2 & \small{Input + Output} &
14.25 &
32.7 &
54.4 &
10.10 &
49.3 &
62.4
\\
LLVQ (spherical shaping, Spherical GPTQ) (\textbf{ours}) & Yes & 2 & \small{Input + Output} &
14.18 &
28.6 &
55.1 &
10.10 &
50.2 &
\textbf{63.0}
\\
LLVQ (shape-gain w/ 0 bit gain, Euclidean GPTQ) (\textbf{ours}) & Yes & 2 & \small{Input + Output} &
16.92 &
26.1 &
51.4 &
11.01 &
43.4 &
58.9
\\
LLVQ (shape-gain w/ 0 bit gain, Spherical GPTQ) (\textbf{ours}) & Yes & 2 & \small{Input + Output} &
\textbf{13.66} &
\textbf{34.5} &
\textbf{55.3} &

\textbf{9.73} &
\textbf{51.3} &
\textbf{62.9}
\\
LLVQ (shape-gain w/ 2 bit gain, Euclidean GPTQ) (\textbf{ours}) & Yes & 2 & \small{Input + Output} &
14.94 &
27.6 &
53.6 &

10.40 &
48.9 &
61.9
\\
LLVQ (shape-gain w/ 2 bit gain, Spherical GPTQ) (\textbf{ours}) & Yes & 2 & \small{Input + Output} &
14.42 &
33.3 &
54.4 &
10.17 & 49.2 & 62.9
\\
\hline
\end{tabular}
}
\vspace{-0.3em}
\end{table*}

\end{document}